\documentclass[11pt]{article}

\usepackage{acl}
\usepackage[utf8]{inputenc}

\usepackage{times}
\usepackage{latexsym}
\usepackage{graphicx}
\usepackage{multicol}
\usepackage{multirow}
\usepackage{booktabs}
\usepackage{amsmath,amssymb}
\usepackage{subcaption}
\usepackage{soul}
\usepackage{lipsum}  

\usepackage[LAE,T1]{fontenc}
\usepackage[utf8]{inputenc}
\usepackage{amsfonts}
\usepackage{dsfont}
\usepackage{microtype}
\usepackage[arabic,english]{babel}
\title{An Analysis of Multilingual FActScore}

\author{Vu Trong Kim$^1$, Michael Krumdick$^2$, Varshini Reddy$^2$ \\ 
{\bf Franck Dernoncourt$^3$, Viet Dac Lai$^2$} \\
   $^1$KAIST  $^2$Kensho Technologies   $^3$Adobe Research\\
  \texttt{kim\_vu\_010801@kaist.ac.kr} \;\;\;
  \texttt{franck.dernoncourt@adobe.com}  \\ \texttt{\{michael.krumdick,varshini.reddy,viet.lai\}@kensho.com}
}

\newcommand{\red}[1]{\textcolor{red}{#1}}
\newcommand{\mtc}[2]{\multicolumn{#1}{|l|}{#2}}
\newcommand{\mtr}[2]{\multirow{#1}{*}{#2}}

\newcommand{\langar}[1]{``\foreignlanguage{arabic}{#1}''}

\begin{document}

\maketitle

\begin{abstract}

FActScore has gained popularity as a metric to estimate the factuality of long-form texts generated by Large Language Models (LLMs) in English. However, there has not been any work in studying the behavior of FActScore in other languages. This paper studies the limitations of each component in the four-component pipeline of FActScore in the multilingual setting. We introduce a new dataset for FActScore on texts generated by strong multilingual LLMs. Our evaluation shows that LLMs exhibit distinct behaviors in both fact extraction and fact scoring tasks. No LLM produces consistent and reliable FActScore across languages with varying levels of resources. We also find that the knowledge source plays an important role in the quality of the estimated FActScore. Using Wikipedia as the knowledge source may hinder the true FActScore of long-form text due to its limited coverage in medium- and low-resource languages. We also incorporate three mitigations to our knowledge source that ultimately improve FActScore estimation across all languages. 
% \footnote{\url{http://github.com/anonymous}}

\end{abstract}

\section{Introduction}

Recent advancements in LLMs have demonstrated significant capabilities \cite{brown2020language, chowdhery2022palm, anil2023palm, geminiteam2024gemini, openai2024gpt4} in many applications \cite{zhao2023survey}.
Despite this advancement, LLMs remain prone to generate false information in response to information-seeking queries \cite{huang2023survey,min-etal-2023-FActScore}. 
% These false responses contain information that conflicts with established knowledge such as incorrect dates of birth or occupations of well-known figures, hence, undermining the reliability and broader application of these models. 
To address this critical problem, LLMs have been trained at unprecedented scales \cite{brown2020language, chowdhery2022palm} to cope with the massive world knowledge and aligned to reduce hallucination \cite{shi2024incontext,chuang2024dola,dhuliawala2023chainofverification}. To further prevent the generation of false information, the Retrieval Augmented Generation method provides retrieved documents from trustworthy sources to the LLM \cite{ram2023context,yu2023improving}. 

FActScore was introduced to estimate the factuality of generated texts automatically \cite{min-etal-2023-FActScore} and at a low cost by combining LLM-as-a-judge scoring \cite{zheng2024judging} with existing reliable knowledge sources such as Wikipedia. FActScore has been enhanced to incorporate a larger knowledge base, like the internet, and to utilize more powerful retrieval models such as Google Search, resulting in better estimation across a larger domain coverage \cite{wei2024long}.

With the rapid development of multilingual LLMs \cite{ai2024yi,aryabumi2024aya}, many more people are interacting with LLMs in an increasingly diverse set of languages. Hence, there is a crucial need to monitor and improve the factuality of texts beyond just the English language, making it helpful and safe for users across the entire world \cite{huang2023survey, ji2023mitigating}.

In this paper, we study the feasibility of the FActScore pipeline \cite{min-etal-2023-FActScore} in a multilingual setting.
The FActScore pipeline consists of multiple components: a knowledge source, a retrieval model, an LLM-based fact extractor, and an LLM-based fact scorer. We aim to scrutinize each component individually to identify bottlenecks and address these issues. However, there is no existing multilingual dataset for evaluating FActScore besides the original English-only dataset published by \citet{min-etal-2023-FActScore}.
To bridge this gap, we annotate a new {\bf native} dataset of factuality in 3 non-English languages representing high-, medium-, and low-resource levels. This dataset is created on the texts generated by strong multilingual LLMs, i.e., GPT-4 and Gemini-Pro-1.0. We find that all evaluator models show decreased FActScore accuracy in lower-resource languages.
We attribute this to several components. First, the performance of fact extraction, the simplest task in the FActScore pipeline, deteriorates with lower resource languages. To address this issue, we finetuned an open-source LLM for this task and achieved better performance than GPT-3.5. Second, the quality of the knowledge source is crucial to the overall accuracy of FActScore. Higher resource languages typically have Wikipedia pages with higher quality and coverage, leading to better FActScore estimation.
Using the Internet as the knowledge source \cite{wei2024long}, therefore, has the greatest impact on improving the accuracy of FActScore estimation in medium and low-resource languages.

Our contributions are as follows:

\begin{itemize}
    \item We annotated a new {\bf native} dataset on the text generated by 2 strong multilingual LLMs in 3 languages for the multilingual FActScore task.
    \item We highlighted the importance of selecting knowledge sources in evaluating FActScore in the multilingual setting due to the variation in the quality of the knowledge sources in different languages.
    \item We found that increasing the quality of the knowledge source, either by utilizing the Internet or even another LLM's internal knowledge, has a great impact in improving the FActScore accuracy in all languages.
\end{itemize}

\section{Related Work}

With the advancement of language model development, numerous methods have been proposed to assess their factual alignment. 
A significant portion of these efforts involves using questions and corresponding short answers  \cite{lin2021truthfulqa,li2023halueval}, slot-filling \cite{cheng2023evaluating} task related to specific pre-collected factoids, however, they do not reflect practical use cases \cite{huang2023survey}. 
Instead, directly assessing open-ended generated texts offers a clearer signal of the level of factuality in real use cases \cite{huang2023survey}. 
\citet{min-etal-2023-FActScore} estimate the FActScore of biographies generated by LLMs by evaluating individual candidate facts in the text.
\citet{wei2024long} extended topic coverage and utilized the Google API to query references for evaluation, thereby accessing a broader range of domains.
Our study builds heavily on these approaches, focusing on the effectiveness of FActScore across high-, medium- and low-resource languages.
In these scenarios, both the language models' performance in each component of the evaluation pipeline and their multilingual capabilities are critical. 
Other approaches rely on language models’ internal knowledge pools for factuality assessment \cite{azaria2023internal,dhuliawala2023chainofverification}. 
While this approach offers simplicity, it raises concerns about the intrinsic factual alignment of these evaluators.

Considering multilingual factuality, X-FACTR \cite{jiang2020x} and MLAMA \cite{kassner-etal-2021-multilingual}, adapted from LAMA \cite{petroni2019language}, assess models’ relational knowledge through the “fill-in-the-blank” task. X-Fact \cite{gupta2021x} releases a multilingual fact-checking benchmark, a factual correctness classification task covering various topics and 25 typologically diverse languages across 11 language families. \citet{qi2023cross} introduces an extension of MLAMA and X-FACTR and a new metric to assess the cross-lingual consistency of language models. While these attempts shed light on multilingual factuality alignment, they mainly involve pre-collected sets of factual statements. Our work aims to evaluate the factuality of open-ended text generation.

\citet{shafayat2024multi} adapted FActScore for a multilingual context by translating the biographies to English. Our work investigates both translation and performing the entire FActScore pipeline directly in the reference language. We also designed a comprehensive set of biographies to better capture the cultural proclivities of the target population.

\section{Tasks \& Resources}

In this work, we evaluate the FActScore in multilingual settings using two resources: a translated annotation from previous work and a new native annotation.

\subsection{Tasks}

\newcommand{\bios}[0]{\mathcal{X}}
\newcommand{\extr}[0]{\mathcal{E}}
\newcommand{\subj}[0]{\mathcal{M}}
\newcommand{\ksrc}[0]{\mathcal{C}}
\newcommand{\eval}[0]{\mathcal{V}}

The FActScore pipeline \cite{min-etal-2023-FActScore} consists of two main steps: 

{\bf Atomic Fact Extraction} that employs an extractor $\extr$ to break a long-form biography $x$ generated by a subject LLM $\subj$ into atomic candidate facts $A^{\extr}(x)= \{a^{\extr,x}_i\}$

{\bf Factuality Scoring}  that employs an evaluator $\eval$ assigning a binary (\textit{supported/not supported}) label $y^{\extr,x,\eval,\ksrc}_i$  to every candidate fact $a_i$ based on a knowledge source $\ksrc$.

The final FActScore estimates the precision of the generated biographies $\bios$:
$$
f_{\ksrc,\eval}(\extr,x) = \frac{1}{|A^{\extr}(x)|}\sum_{a_i \in A^{\extr}(x)}{\mathds{1}(a_i)}
$$
$$
\text{FActScore}_{\extr,\ksrc,\eval}(\subj)=\mathbb{E}_{x\in \bios}[f_{\ksrc,\eval}(\extr,x)]
$$

\subsection{Translated Annotation (en $\rightarrow$ X) (R1)}

The original FActScore published a set of biographies $\bios^\subj$ generated by several subject LLMs $\mathcal{M}$ and their corresponding FActScore \cite{min-etal-2023-FActScore} with full annotation of atomic fact and supporting label pairs $(a^{\extr,x}_i, y^{\extr,x,\eval,\ksrc}_i )$. We use Google Translate to translate each atomic fact $a^{\extr,x}_i$ in English into every other target language $t$ to produce a new translated annotation $(a^{\extr,x, t}_i, y^{\extr,x,\eval,\ksrc}_i )$. The knowledge source $\ksrc$ (written in English) is also translated into corresponding target languages. We select a set of target languages (X) in 3 groups: high-resource (i.e., French (fr), Spanish (es), Chinese (zh-cn), Russian (ru), and Vietnamese (vi)), medium-resource (i.e., Arabic (ar) and Hindi (hi)), and low-resource (i.e., Bengali (bn)). 

\subsection{Native Annotation (R2)}

The translated annotations are able to provide some insights into potential issues with FActScore in the multilingual setting. However, they provide a confounding factor: cascading errors due to issues with the translations themselves.
This is especially relevant for low-resource languages. Therefore, we also annotate new FActScore data in non-English languages to better estimate FActScore and explore the issues of this task. In particular, we aim for a broad language coverage spanning high-, medium-, and low-resource languages. We investigated one language across each of these resource categories: Spanish, Arabic, and Bengali, respectively.

Following \citet{min-etal-2023-FActScore}, we carefully curated a set of biographies for each language are from 4 geographical regions and 5 levels of rarity (See Appendix \ref{app:biography_geo}). We attempted to use the same generative models as in \cite{min-etal-2023-FActScore}. However, these models are not explicitly designed to be multilingual and as a result, could not generate biographies of an acceptable quality, specifically in the low-resource language. To address this, we analyze the performance of explicitly multilingual LLMs, i.e., GPT-4 (GPT4) and Gemini Pro  (GemP) to generate biographies. 

We hired 2 native annotators for each language and followed the same annotation guidelines by \citet{min-etal-2023-FActScore} to evaluate the true FActScore of generated text. 
The Kappa agreement scores between Spanish, Arabic, and Bengali annotators are 79.8, 73.1, and 80.2, respectively. These show a substantial agreement (61-81) to close to almost perfect agreement (81-100) between native annotators.

% \begin{table*}
%     \begin{tabular}{lllllllllll}
%         Lang & Subject model & \# of bios & \# of relevant & \# of irrelevant & \# of abstain & relevant percentage & irrelevant percentage & abstain percentage & \# of facts from relevant bios & \# facts per relevant bio\\
%         Spanish & Gemini-Pro & 94 & 56 & 27 & 11 & 59.57 & 28.72 & 11.70 & 4486 & 80.11\\
%         Spanish & GPT4-Turbo & 95 & 67 & 1  & 27 & 70.53 & 01.05 & 28.42 & 5468 & 81.61\\
%         Arabic  & Gemini-Pro & 95 & 57 & 36 & 2  & 60.00 & 37.89 & 2.11  & 4065 & 71.32\\
%         Arabic  & GPT4-Turbo & 95 & 59 & 8  & 28 & 62.11 & 08.42 & 29.47 & 3650 & 61.86\\
%         Bengali & Gemini-Pro & 94 & 54 & 40 & 0  & 57.45 & 42.55 & 0.00  & 3327 & 61.61\\
%         Bengali & GPT4-Turbo & 86 & 54 & 29 & 3  & 62.79 & 33.72 & 3.49  & 2492 & 46.15\\
%     \end{tabular}
% \end{table*}

\begin{table}[!h]
\resizebox{\linewidth}{!}{
    \begin{tabular}{llcrrrccc}
    \toprule
        & \mtr{2}{\bf Subject} & \mtr{2}{\bf \#Bios} & \mtr{2}{\bf R} & \mtr{2}{\bf I} & \mtr{2}{\bf A} & \mtr{2}{\bf \#Facts} & \multicolumn{2}{c}{\bf FActScore}\\
        \cline{8-9}
        & & & & & & & \bf WN-1 & \bf WN-All\\
        \midrule
        \mtr{2}{es}     & GemP & 100 & 62 & 27 & 11 & 79.4 & 67.20 & 70.77\\
                        & GPT4 & 100 & 72 &  1 & 27 & 81.4 & 82.86 & 86.83\\
        \midrule
        \mtr{2}{ar}     & GemP & 100 & 61 & 37 &  2 & 70.9 & 59.27 & 61.81\\
                        & GPT4 & 100 & 63 &  8 & 29 & 61.8 & 74.34 & 78.10\\
        \midrule
        \mtr{2}{bn}     & GemP & 100 & 60 & 40 &  0 & 58.9 & 58.77 & 60.55\\
                        & GPT4 & 100 & 68 & 29 & 3  & 46.4 & 71.95 & 74.48\\
        \bottomrule
    \end{tabular}
}
\caption{Statistics of the generated biographies by {\bf GemP} and {\bf GPT4} including the percentage of Relevant ({\bf R}), Irrelevant ({\bf I}), Abstain ({\bf A}) biographies; the average number of atomic facts in relevant generated biographies; and FActScore evaluated by native speakers using one native Wikipedia page (WN-1) and the whole native Wikipedia (WN-All). Note that the FActScore is computed on the relevant generated texts only.}
\label{tab:biography_generation}
\end{table}

Table \ref{tab:biography_generation} presents the statistics of generated biographies by both subject models. 
Both models generate more candidate atomic facts in higher-resource languages than lower-resource languages, however, this phenomenon seems to be clearer with GPT4. GPT4 generates more relevant biographies than GemP in all three languages. GPT4 also abstains significantly more than GemP in Spanish and Arabic, whereas GemP produces many more irrelevant biographies. This shows that GPT4 has a broader knowledge and a higher awareness of its knowledge limitation.
In terms of FActScore, GPT4 yields much higher FActScore(s) than GemP in all three languages, using either a single Wikipedia page or the whole Wikipedia with an average margin of approximately 14.6\%. Last but not least, FActScore(s) evaluated based on the whole Wikipedia (WN-All) are higher than FActScore evaluated on a single Wikipedia page (WN-1) in all cases (on average 3\%). This suggests that a larger knowledge source gives a higher FActScore. In other words, the knowledge source is the ceiling of evaluating factuality.

\section{Experiments}
\subsection{Atomic Fact Extraction}
\label{sec:fact_extraction}
FActScore decomposes a long-form text into multiple atomic statements, each containing a single piece of information. 
The original methodology uses few-shot demonstrations to prompt InstructGPT for this task \cite{min-etal-2023-FActScore}. 
We examine the performance of different models and pinpoint issues of existing models for this task.

{\bf Settings:} Due to the higher quality of text generated in English, prior work by \citet{min-etal-2023-FActScore} only considered if the candidate facts need to be merged or split, mainly concerning whether the facts are atomic. However, in a multilingual setting, the texts generated by LLM may contain other kinds of errors where the facts need to be merged or split, not grounded, duplicated, missing some information, and linguistic errors.

We choose GPT-3.5 (GPT3.5), GPT4, and Gemma for evaluation in this task. These models were selected for their best performance via a pilot study on a small subset of R1 (See Appendix \ref{app:fact_extraction_pilot}). We evaluate the GPT3.5 and GPT4 as few-shot In-Context Learning while Gemma is further supervised finetuned for this task. In particular, we finetune Gemma on 42k pairs of (sentence, extracted atomic facts) derived from R1. Then these three models are evaluated on a subset of 200 sentences, sampled randomly from R2 with a 1:1 ratio facts generated by GPT4 and GemP.

{\bf Results:} Table \ref{tab:fact_extractor} shows the number of errors made by 3 models (GPT3.5, Finetuned Gemma, and GPT4).
Among these three models, GPT4 is the best model by a relatively large margin across all three languages.
Finetuned Gemma is competitive to GPT3.5 in high-resource and better in low-resource and medium-resource languages.

However, GPT4 and GPT3.5 's performances deteriorate rapidly with low-resource language (approximately double the average error rate in Bengali, compared to Spanish and Arabic). On the other hand, the FT Gemma does not show a performance reduction in low-resource language. In fact, its error rate in Bengali is lower than those in Spanish and Arabic. This suggests that finetuning has potentially helped this model maintain a steady performance across all resource languages.

More importantly, due to the better performance of LLMs in English, \citet{min-etal-2023-FActScore} did not consider other types of errors that may happen in multilingual settings.
In particular, we see a large number of grounding errors in medium and low-resource languages (Arabic and Bengali), while we don't see that in high-resource languages such as Spanish. 
LLMs also missed some detailed information in the given generated text in this task.

\begin{table}[t]
\resizebox{\linewidth}{!}{
    \begin{tabular}{llcrrrrrrc}
    \toprule
        \bf & \bf Extractor & \bf \#Sent & \bf M & \bf S & \bf G  & \bf D & \bf I & \bf L & \bf Avg. $\downarrow$\\
        \midrule
        \mtr{3}{es} & GPT3.5 & 192 & 1 & 24 & 0 & 9 & 14 & 0 & 0.25\\
                     & FT Gemma & 192 & 0 & 35 & 0 & 0 & 14 & 0 & 0.26\\
                     & GPT4 & 192 & 1 & 2 & 0 & 3 & 9 & 0 & \bf 0.08\\
        \midrule
        \mtr{3}{ar}  & GPT3.5 & 180 & 1 & 22 & 10 & 2 & 13 & 3 & 0.28\\
                     & FT Gemma & 180 & 0 & 20 & 10 & 0 & 14 & 2 & 0.26\\
                     & GPT4 & 180 & 3 & 0 & 5 & 0 & 8 & 0 & \bf 0.09\\
        \midrule
        \mtr{3}{bn} & GPT3.5 & 175 & 3 & 36 & 19 & 12 & 24 & 8 & 0.58\\
                     & FT Gemma & 175 & 2 & 22 & 5 & 2 & 7 & 1 & 0.22\\
                     & GPT4 & 175 & 2 & 9 & 3 & 2 & 8 & 0 & \bf 0.14\\
        \bottomrule
    \end{tabular}
}
\caption{Fact Extraction: Total number of errors by categories (Need Merge (M), Need Split, Not Grounded (G), Duplicated (D), Missing Information (I), and Linguistic Error (L)), and the average number of errors per sentence on texts generated by GPT4 and GemP .}
\label{tab:fact_extractor}
\end{table}

\subsection{Factuality Scoring}
\label{sec:factuality_scoring}

This section investigates the feasibility of using LLMs as factuality scorers in multilingual settings.

{\bf Settings: } We use GPT4 to extract facts from biographies generated by two subject models namely GPT4 and GemP to provide the same denominator for this evaluation. We evaluate 4 LLMs as fact scorers (GPT3.5, GPT4, Mistral, and GemP) on the text generated by GPT4 and GemP in native languages. The human-annotated dataset (R2) is used as the ground truth.

\begin{figure}[t]
    \begin{subfigure}{\linewidth}
        \includegraphics[width=\linewidth]{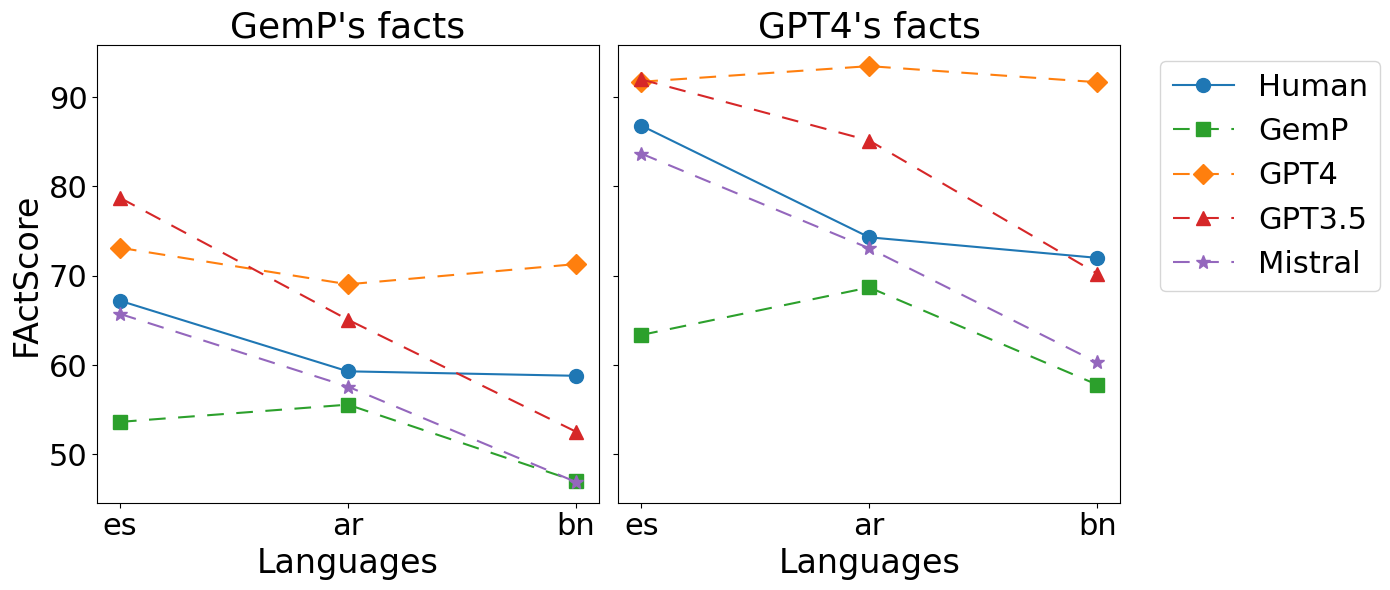}
    \end{subfigure}
    \begin{subfigure}{\linewidth}
        \includegraphics[width=\linewidth]{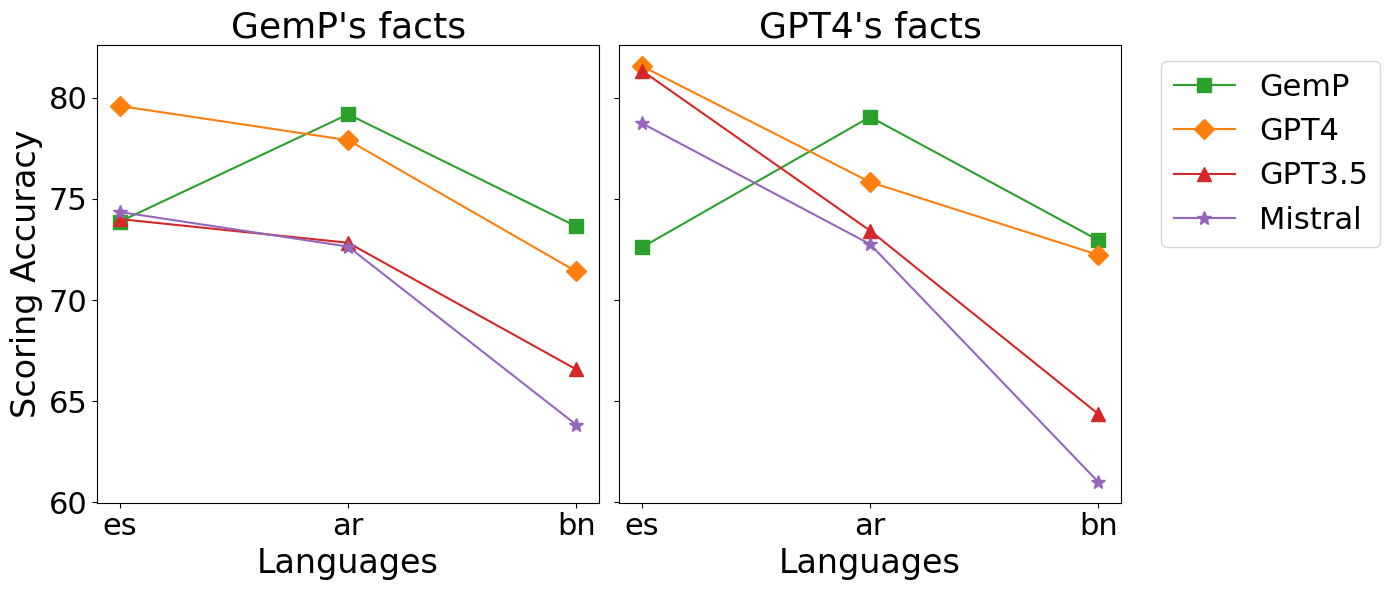}
    \end{subfigure}
    \caption{FActScore (upper) and Scoring Accuracy (lower) predicted by 4 scorers (GPT4, GemP, GPT3.5, Mistral) in comparison with FActScore by human (R2) on texts generated by GPT4 and GemP in native languages.}
    \label{fig:FActScore_on_r2}
\end{figure}

{\bf Results:} Figure \ref{fig:FActScore_on_r2} (upper) shows the FActScore predicted by LLMs and by humans (R2).
GemP consistently underestimates FActScore, whereas GPT4 significantly overestimates FActScore across both subject models. GPT3.5 overestimates Spanish and Arabic while closely estimating FActScore for Bengali. On the other hand, Mistral closely estimates FActScore for Spanish and Arabic while substantially underestimating the FActScore for Bengali.
This experiment suggests that none of these models offers a reliable FActScore across the whole spectrum of languages, even with strong LLMs (e.g., GPT4 and GemP).

Figure \ref{fig:FActScore_on_r2} (lower) shows the scoring accuracy of the LLM scorers. GemP shows a steady accuracy on both GPT4 and GemP facts. Its accuracy does not show a clear dependency on the resource level. On the other hand, the accuracy of GPT4, GPT3.5, and Mistral decreases in turn with the level of language resources. In particular, GPT3.5 and Mistral's accuracy decreases at a steeper pace than GPT4's. Further discussion on this component will be provided in Section \ref{sec:dicussion}.
% The above replaces the below
% The reason for this is further discussed in Section \ref{sec:qualitative_analysis}.

\subsection{Knowledge Source}
\label{sec:knowledge_source}
Since FActScore is a function of knowledge source \cite{min-etal-2023-FActScore}, the quantity and quality of the information of the knowledge source greatly affect the subsequent score \cite{wei2024long}. 
This section investigates the sensitivity of FActScore to changes in the underlying knowledge sources. 

{\bf Settings:} We collected 32 biographies of entities per language in four categories of popularity and geographical relevance: {\it internationally popular}, {\it internationally unpopular}, {\it locally popular}, and {\it locally unpopular} (See Appendix \ref{app:biography_geo}).
The annotators evaluate facts using three different sources: 
the native Wikipedia, the English Wikipedia, and the whole Internet.
Since the Internet is a superset of knowledge sources, we considered the annotations created with access to the Internet as the golden annotations for evaluating the quality of other knowledge sources.

\begin{figure}
    \centering
    \includegraphics[width=\linewidth]{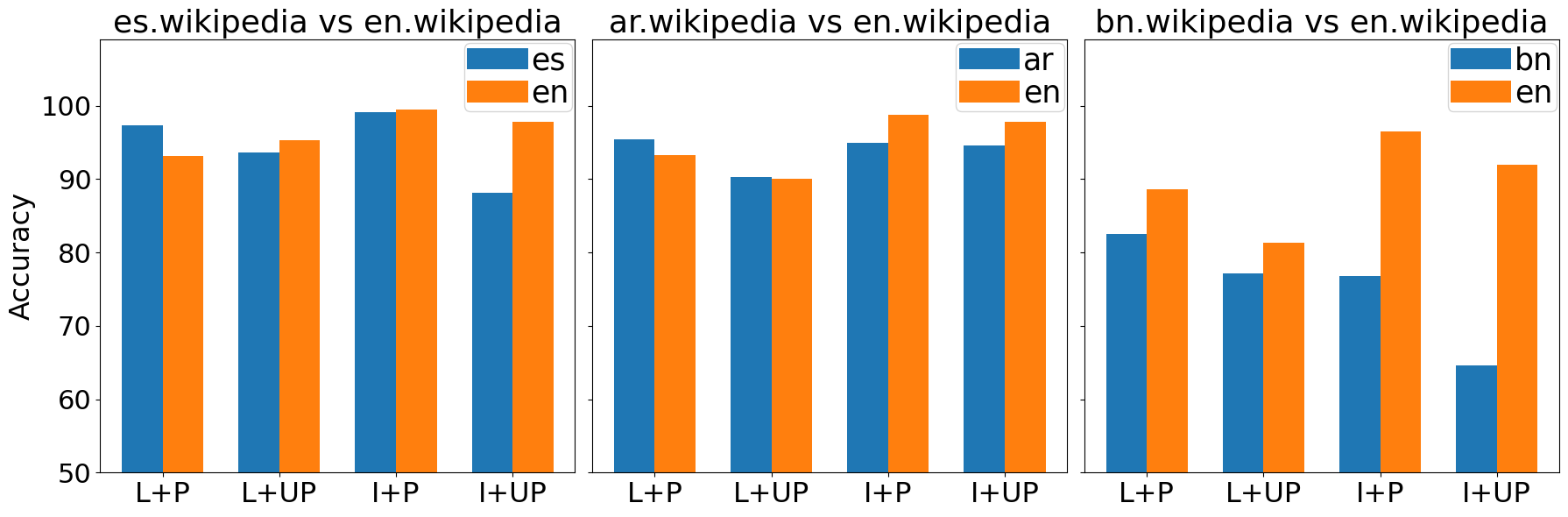} 
    \caption{Accuracy of Factuality Scoring task with different knowledge sources. {\bf L} stands for Local/Domestic, while {\bf I} stands for International. {\bf P} stands for Popular and {\bf UP} stands for UnPopular.}
    \label{fig:knowledge_source}
\end{figure}

{\bf Results:} Figure \ref{fig:knowledge_source} shows the scoring accuracy between evaluating 4 categories of popularity in 3 languages. 
Using Spanish Wikipedia pages yields higher accuracy in labeling locally popular figures (L+P), whereas English Wikipedia pages are better for internationally unpopular entities (I+UP).
For Arabic, the Arabic Wikipedia is better for local popular entities (L+P), while the English Wikipedia is better for international entities (I+P and I+UP). 
For Bengali, the Bengali Wikipedia has a much lower performance compared to the English counterpart in all four categories, especially for the international entities (I). This suggests that Bengali Wikipedia has a very low coverage, inadequate for most cases.
Last but not least, even though English pages provide better coverage for local entities (L+P and L+UP) than Bengali pages, the scoring accuracies using English pages for Bengali local entities are still lower than those of international entities. These differences in performance between international and local figures highlight the importance of choosing local entities and local knowledge sources in multilingual FActScore evaluation and estimation.

\subsection{Retriever}
\label{sec:retriever}
Due to the limitation of the LLM context length, a Wikipedia page of the evaluated entity is split into short passages. A retriever model retrieves $k$ relevant passages. These passages are used as reference knowledge sources.

{\bf Settings:} We use a multilingual version of SentenceBERT \cite{reimers2019sentencebert} as the retriever model instead of the English-only retriever, T5 \cite{ni2021large}, used in the original work.
We use R1 for this experiment because it also provides the ranking of the retrieved passages. For each translated fact, $k=5$ retrieved passages are retrieved out of all passages. We measure the Recall@k and the average hit rate of the top 1 and top 2 passages.

{\bf Results:}
Table \ref{tab:retrieval_assessment} reports the retrieval performance of the retrieval models on 9 languages of 3 resource-level groups. There is a notable decrease in match rates for lower-resource languages. However, for high- and medium-resource languages, a retrieval match of over 60 percent is observed, equivalent to 3 out of 5 passages retrieved by the English retriever also being retrieved by the multilingual retriever. Conversely, in Bengali, a low-resource language, the Recall@5 drops significantly to just 50\%.
We see a similar pattern in the hit rate of the Top 1 and Top 2 passages.
The effect of the retrieval model is further discussed in Section \ref{sec:increase_no_passages}. Increasing the number of retrieved passages and leveraging the longer context length of recent language models \cite{xiong2023effective}, can potentially mitigate this issue and significantly boost the accuracy of the pipeline.

\begin{table}[t]
\centering
\resizebox{0.8\linewidth}{!}{
    \small
    \begin{tabular}{lcccc}
    \toprule
        \bf Resource & \bf Lang & \bf Recall@5 & \bf Top 1 & \bf Top 2\\
        \midrule
        \mtr{6}{High}
        & en    & 67.7 & 85.8 & 94.6\\
        & fr    & 66.0 & 84.4 & 93.9\\
        & es    & 66.1 & 84.8 & 94.3\\
        & ru    & 65.0 & 82.8 & 92.4\\
        & zh    & 64.5 & 82.8 & 92.7\\
        & vi    & 65.4 & 83.6 & 93.2\\
        \midrule
        \mtr{2}{Medium} 
        & ar    & 63.3 & 81.4 & 91.7\\
        & hi    & 61.5 & 78.3 & 89.6\\
        \midrule
        \mtr{1}{Low} 
        & bn    & 51.2 & 51.8 & 69.6\\
        \bottomrule
    \end{tabular}
}
\caption{Retrieval performance of the multilingual retriever in Recall@k (\%). Top 1 and Top 2 measure the average hit rate (HR@5) of retrieving the original Rank 1 and Rank 2 passages.}
\label{tab:retrieval_assessment}
\end{table}

\section{Discussion}
\label{sec:dicussion}
\subsection{Would translation help?}

\label{sec:transaltion}

A simple method for a multilingual FActScore is first translating non-English long-form text and knowledge sources into English ({\bf X$\rightarrow$en}) and estimating the FActScore on these proxy translated English texts \cite{shafayat2024multi}.
This is a promising method given that the quality of machine translation has improved significantly in the last decade.
To do this, we translated the Native Annotation (R2) into English to get a translated English annotation (R3).

\begin{figure}[t]
    \centering
    \includegraphics[width=\linewidth]{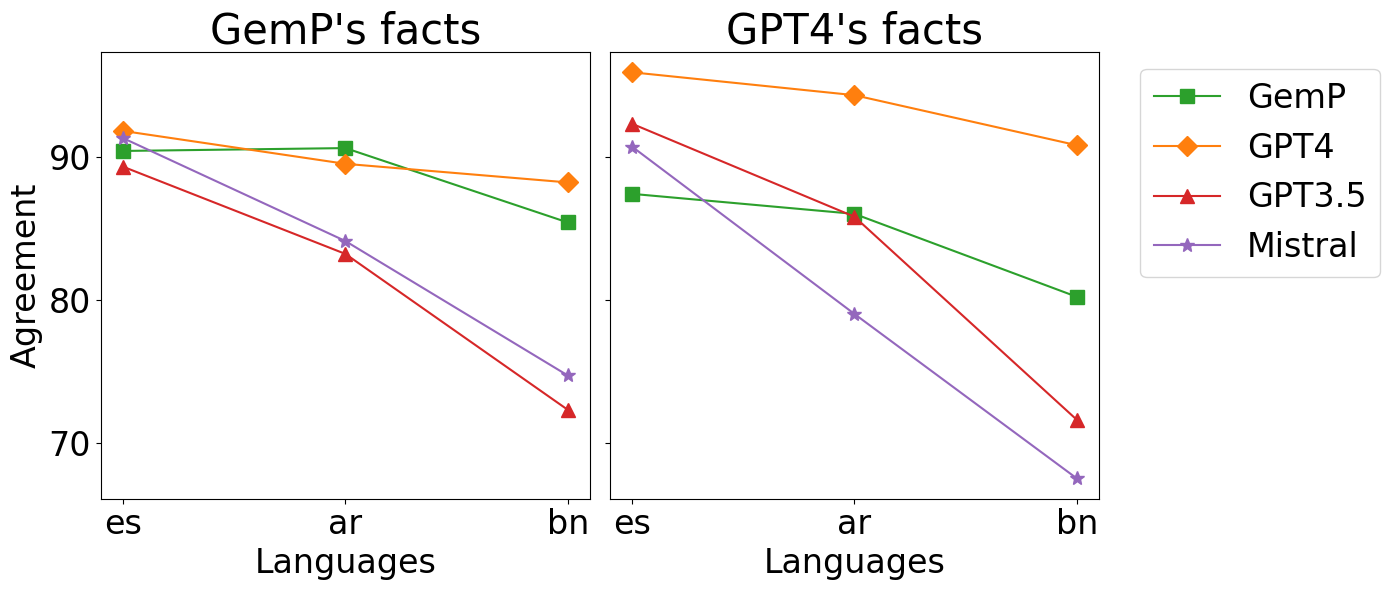}
    \caption{Prediction agreement between two variants of facts (in target language and in translated English).}
    \label{fig:cross_lingual}
\end{figure}

Figure \ref{fig:cross_lingual} shows the prediction matching for the Factuality Scoring task on texts in the target language and in translated English. GemP and GPT4 are the two strong scorers with consistently high matching, GPT3.5 and Mistral have significantly lower matching scores in lower-resource languages. Additionally, GPT4 and GemP see a slighter decline in matching scores for lower-resource languages than GPT3.5 and Mistral. This matching variation across different languages for this task among even the most advanced LLMs may lead to unreliable FActScore estimation in lower-resource languages.

\begin{figure}[t]
    \begin{subfigure}{\linewidth}
        \includegraphics[width=\linewidth]{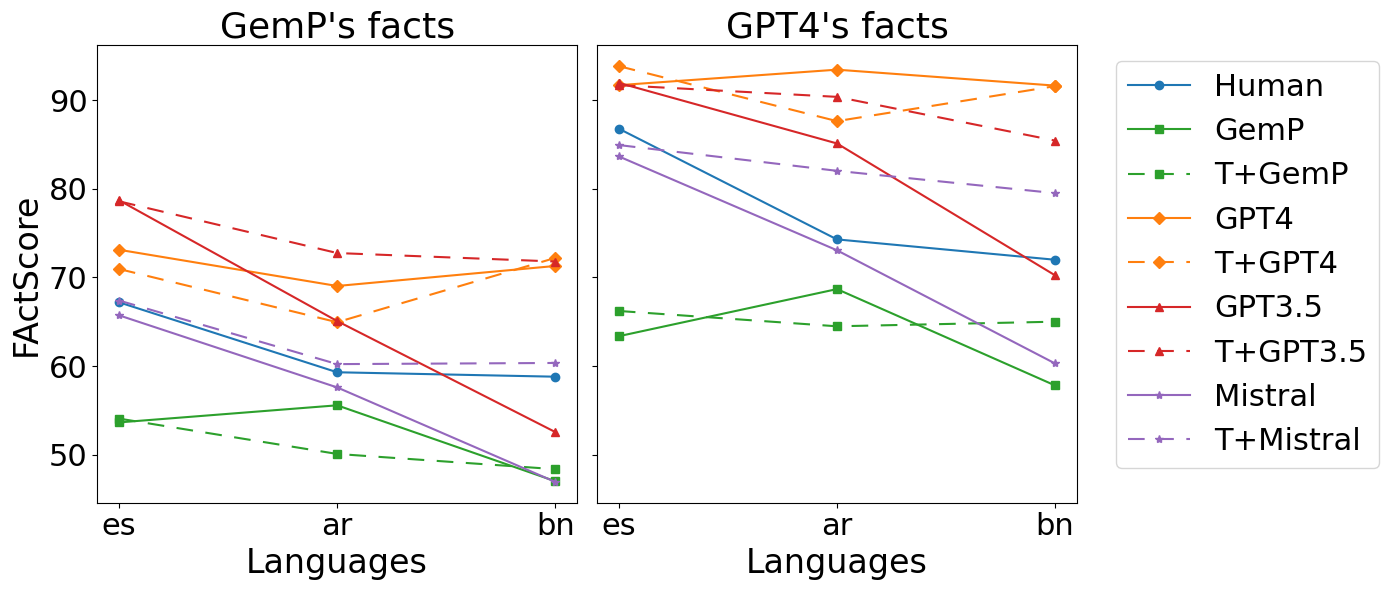}
    \end{subfigure}
    \begin{subfigure}{\linewidth}
        \includegraphics[width=\linewidth]{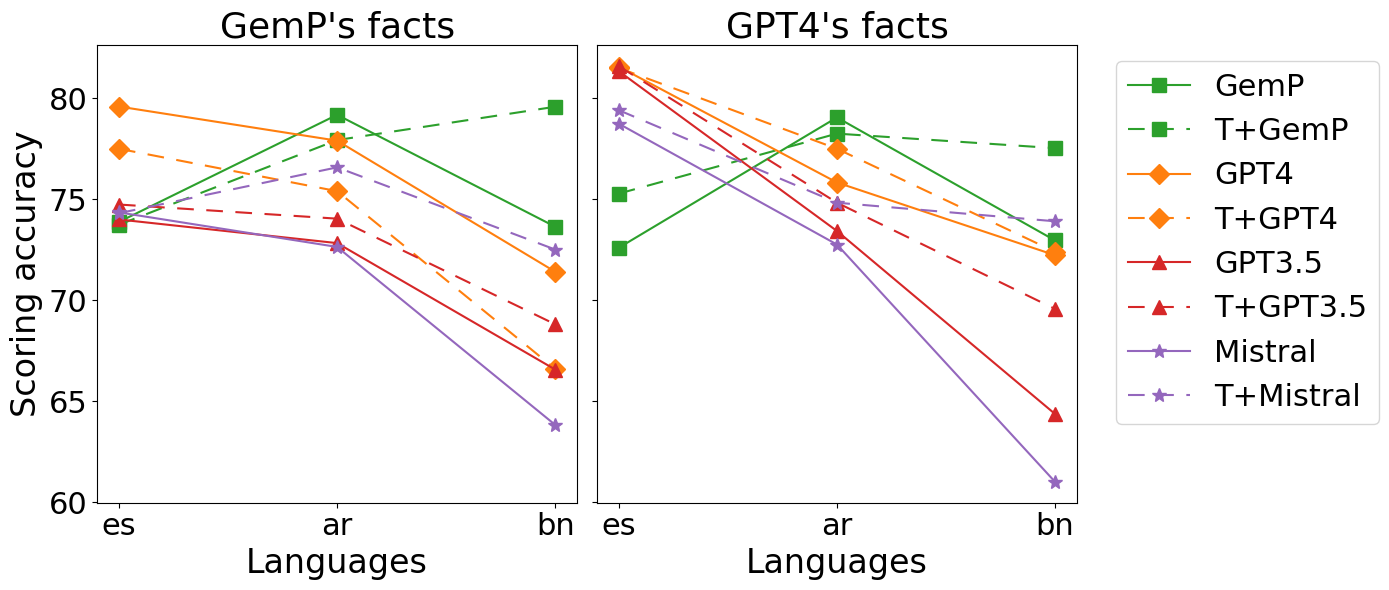}
    \end{subfigure}
    \caption{FActScore (upper) and Scoring accuracy (lower) by fact scorers with and without translation in comparison with FActScore by human (R2) on texts generated by GPT4 and GemP. Dash lines denote the translation being used, along with corresponding scorers.}
    \label{fig:fact_score_on_r3}
\end{figure}

Figure \ref{fig:fact_score_on_r3} (lower) compares the scoring accuracy between using R3 and using R2. We see a significant improvement in scoring accuracy for Mistral and GPT3.5 in Arabic and Bengali and GemP in Bengali, all on both GPT4 and GemP texts. We attributed this to both better reading comprehension and retrieval performance in English compared to non-English languages, especially Bengali. Appendix \ref{app:impact_translation_retriever} explores the impact of translation on retrieval performance in more detail.
On the other hand, we see a significant decline in the accuracy of the scorer GPT4 on GemP's texts for all three languages while a slight increase in the accuracy in Arabic on GPT4 texts. 

Figure \ref{fig:fact_score_on_r3} (upper) shows the FActScore predicted by these models in native texts and translated English texts.  The translation contributes to the overestimation of FActScore by GPT3.5 and Mistral in medium and low-resource languages. On the other hand, translation has little effect on stronger scorers such as GPT4 and GemP. This suggests that these models are more consistent in understanding both English and non-English texts.

\subsection{Error analysis}
\label{sec:qualitative_analysis}

Figure \ref{fig:fact_score_on_r3} shows significant differences in the factuality-scoring task remain between the most advanced model evaluators, i.e., GPT4 and GemP, and native speakers. We conducted an error analysis to investigate the categories of these disagreements. For each language and each subject model, we randomly select 60 disagreement samples between LLMs and humans. We manually inspect this to identify the primary disagreement.

Table \ref{tab:error_analysis} reports the raw number of errors. The primary cause of errors by the scorer GPT4 is contextual unfaithfulness, accounting for 73\% of the errors across 3 languages and 2 subject models. This issue is more severe in lower-resource languages. However, many contextually unfaithful samples are factually correct according to other knowledge sources beyond the given Wikipedia page. This suggests that GPT4 uses its internal knowledge in the Factuality Scoring task. Appendix \ref{sec:further_analysis} further discusses the behaviors of GPT4 as a scorer. The scorer GemP has a much lower contextual unfaithfulness error (especially factually correct) compared to GPT4. However, GemP makes more errors due to retrieval errors and tabular data. This shows that GemP is more context-dependent and less internal-knowledge-dependent for the Factuality Scoring task. 

\begin{table}[t]
\resizebox{\linewidth}{!}{
\begin{tabular}{llcccccccc}
    \toprule
    & \mtr{2}{\bf Subject} 
    & \multicolumn{3}{c}{\bf Context Unfaithful} 
    & \multicolumn{4}{c}{\bf Others} \\
    \cmidrule(r){3-5} 
    \cmidrule(r){6-9}
    & 
    & \bf FC 
    & \bf Hal. 
    & \bf RD
    & \bf Ret. 
    & \bf Tab.
    & \bf Deb.
    & \bf Misc.\\
    \bottomrule
    \multicolumn{9}{c}{\bf (a) Fact Scorer: GPT4}\\
    \midrule
    \multirow{2}{*}{es} 
        & GemP & 12 & 4 & 2 & 1 & 2 & 4 & 5 \\
        & GPT4 & 17 & 6 & 0 & 1 & 0 & 2 & 4\\
    \cmidrule{2-9}
    \multirow{2}{*}{ar}  
        & GemP & 14 & 5 & 3 & 4 & 1 & 0 & 3\\
        & GPT4 & 17 & 4 & 2 & 0 & 1 & 3 & 3\\
    \cmidrule{2-9}
    \multirow{2}{*}{bn} 
        & GemP & 18 & 1 & 0 & 3 & 2 & 0 & 6\\
        & GPT4 & 22 & 3 & 2 & 0 & 2 & 0 & 1\\
    \bottomrule
    \multicolumn{9}{c}{\bf (b) Fact Scorer: GemP}\\
    \midrule
    \multirow{2}{*}{es} 
        & GemP & 8 & 0 & 5 & 10 & 2 & 3 & 2\\
        & GPT4 & 4 & 0 & 4 & 8 & 6 & 4 & 4\\
    \cmidrule{2-9}
    \multirow{2}{*}{ar}  
        & GemP & 4 & 1 & 7 & 4 & 6 & 1 & 7\\
        & GPT4 & 7 & 1 & 6 & 3 & 6 & 3 & 4\\
    \cmidrule{2-9}
    \multirow{2}{*}{bn} 
        & GemP & 5 & 2 & 5 & 4 & 7 & 3 & 4\\
        & GPT4 & 7 & 2 & 6 & 4 & 5 & 4 & 2\\
    \bottomrule
\end{tabular}
}
\caption{Error analysis: Factually Correct (FC), Hallucination (Hal.), Reading Deficiency (RD)), Retrieval Error (Ret.), Tabular Data (Tab.), Debatable (Deb.), and miscellaneous error (Misc.)) }
\label{tab:error_analysis}
\end{table}

\section{Mitigations}
\label{sec:mitigate_ks_retrieval_limit}

The previous sections have shown evidence of a correlation between lower resource languages, lower retrieval performance (See Table \ref{tab:retrieval_assessment}), lower coverage of the native knowledge source (See Figure \ref{fig:knowledge_source}) and subsequently lower fact scoring accuracy (See Figures \ref{fig:FActScore_on_r2} and \ref{fig:cross_lingual}). To mitigate this problem, we empirically examine three techniques including: improving retrieval performance by (1) increasing the number of retrieved passages, (2) employing language models as Internet search agents and evaluators \cite{wei2024long}, and (3) using language models as a knowledge generator \cite{yu2023generate,chen-etal-2023-beyond}).

{\bf Settings:} We use GemP as the fact scorer for all proposed techniques. 
GemP is more persistent to the change in languages (as shown in Figure \ref{fig:FActScore_on_r2}). It is more sensitive to external knowledge than its internal knowledge (Section \ref{sec:qualitative_analysis}), making it more suitable for evaluating these mitigations than GPT4. The baseline is the original pipeline \cite{min-etal-2023-FActScore} with GemP as the scorer and Wikipedia pages in native languages as the knowledge sources.

We use the 32 generated biographies in the three studied languages that we used to assess knowledge sources in section \ref{sec:knowledge_source}. We consider the facts annotated by native speakers using the whole internet as the golden data. We evaluate these techniques by measuring their scoring accuracy with the golden labels. Table \ref{tab:compare_proposed_method} illustrates the performance of the proposed methods.

\subsection{Expanding Retrieved Passages}
\label{sec:increase_no_passages}
This method increases the number of retrieved passages from 8 to 20, aiming to extend the amount of information given to the scorer. 
This mitigation should alleviate the impact of poor recall in retrieval.
Although the mildest of the three mitigations, this led to a considerable increase in performance across all three languages.
%This suggests significant potential for improving the open-source retriever.
The performance gap is particularly large in Bengali, correlating with observations in Section \ref{sec:retriever} regarding the retriever's deteriorating performance in this language.
This retrieval problem might be further mitigated thanks to the increase in context length of recent language models \cite{xiong2023effective} allowing feeding more information to the LLM-based scorer.

\subsection{Internet as a knowledge source}

\begin{table}[t]
\resizebox{\linewidth}{!}{
    \begin{tabular}{ll|cc|cc}
    \toprule
        & \mtr{2}{\bf Method} & \multicolumn{2}{c|}{\bf FActScore} & \multicolumn{2}{c}{\bf Accuracy}\\
        & & \bf GemP & \bf GPT4 & \bf GemP & \bf GPT4\\
        \midrule
        \mtr{6}{es} 
            % & GemP+Wiki (k=5) & 58.8 & 68.4 & 70.3 & 69.8\\
            & GemP+Wiki (k=8) & 63.3 & 72.8 & 74.3 & 74.0 \\
            & GemP+Wiki (k=20) & 67.0 & 77.3 & 77.3 & 78.5 \\
            & GemP+Google API & 81.5 & 90.3 & 83.2 & 89.9\\
            & GemP+GPT4's IK & 78.8 & 93.5 & 84.6 & 91.9\\
            \cline{2-6}
            & Human+Wiki & 75.3 & 88.4 & 91.9 & 90.1\\
            & Human+Internet & 82.9 & 97.3 & - & -\\
        \midrule
        \mtr{6}{ar} 
            & GemP+Wiki (k=8) & 56.4 & 73.7 & 77.9 & 78.3 \\
            & GemP+Wiki (k=20) & 60.6 & 76.7 & 79.9 & 81.1 \\
            & GemP+Google API & 72.3 & 87.7 & 80.3 & 86.3\\
            & GemP+GPT4's IK & 64.4 & 82.4 & 83.6 & 84.6 \\
            \cline{2-6}
            & Human+Wiki & 60.9 & 81.9 & 90.8 & 89.7\\
            & Human+Internet & 69.2 & 90.5 & - & -\\
        \midrule
        \mtr{6}{bn} 
            & GemP+Wiki (k=8) & 43.8 & 53.0 & 60.6 & 55.1 \\
            & GemP+Wiki (k=20) & 52.6 & 63.5 & 69.1 & 65.7\\
            & GemP+Google API & 75.6 & 88.0 & 86.8& 87.8\\
            & GemP+GPT4's IK & 59.4 & 70.0 & 74.8 & 71.1\\
            \cline{2-6}
            & Human+Wiki & 57.8 & 61.2 & 74.1 & 62.7\\
            & Human+Internet & 82.0 & 97.5 & - & - \\
        \bottomrule
    \end{tabular}
}
\caption{FActScore and accuracy of introduced evaluation methods on GemP and GPT4's generated facts. We use GemP as the LLM scorer. +Wiki (k=x) denotes using x passages from 1 Wikipedia page as references. +Google API denotes using GemP as the Internet search agent and evaluator (evaluation is based on query results). +GPT4's IK denotes using GPT-4's generated Internal Knowledge (IK) and retrieved passages as references. Natives+Wiki/Internet denotes natives, using 1 Wikipedia page or the entire Internet as references for annotations. Natives+Internet is considered as golden labeling to conclude accuracy.}
\label{tab:compare_proposed_method}
\end{table}
Adapted from \citet{wei2024long}, GemP is prompted to send queries to the Google Search API on a given fact and determine the fact's factual accuracy from the query results. We see a clear improvement in fact-scoring accuracy and higher FActScore (closer to the golden) across the subject models and languages. For example, the accuracy on Bengali improved from 60.6 to 86.8.
%and the FActScore improved from 43.8 to 75.6, which is approaching the golden of 82.0, for GemP text in Bengali. 
This shows the benefit of accessing a larger pool of information results in substantial improvement, much greater than merely increasing the number of passages from Wikipedia.

\subsection{LLM as a knowledge source}
\label{sec:llm_gen_k}
Since previous experiments suggested that GPT4 heavily relies on its internal knowledge to assess factuality, we experiment with allowing GPT4 to directly augment the low-coverage knowledge source.
We prompt GPT4 to create a question based on a given fact and then generate related information to answer that question \cite{yu2023generate}. This generated knowledge is combined with retrieved passages, as suggested by \cite{yu2023generate}, and used with a separate evaluator, GemP, for factual labeling. 
It is worth noting that this text is entirely unverified and likely contains some amount of factual errors. 

This approach results in a substantial improvement, larger than that of simply increasing the number of Wikipedia passages across all languages. Compared to using the Google Search API, the GPT4 augmented knowledge base shows higher gains in high- and medium-resource languages. This suggests the reliability of GPT4’s internal knowledge and its effectiveness as a knowledge generator. However, in Bengali, querying evaluation references via Google API yields significantly better factual labeling. 
The improvement from using GPT4's internal knowledge is attributed to the additional relevant information that it provides.

\begin{table}[t]
\centering
\resizebox{\linewidth}{!}{
    \begin{tabular}{ll|cccc|cccc}
    \toprule
        & \mtr{2}{\bf Method} 
        & \multicolumn{4}{c|}{\bf GemP} 
        & \multicolumn{4}{c}{\bf GPT4}\\
        & & \bf TP & \bf FN & \bf FP & \bf TN 
        & \bf TP & \bf FN & \bf FP & \bf TN\\
        \midrule
        \mtr{4}{es} 
            & Wiki (k=8)   & 60.3 & 22.6 & 3.0 & 14.1 & 72.1 & 25.2 & 0.8 & 1.9 \\
            & Wiki (k=20)  & 63.6 & 19.3 & 3.4 & 13.7 & 76.5 & 20.8 & 0.8 & 1.9\\
            & Google API   & 73.8 & 9.1 & 7.7 & 9.4 & 88.7 & 8.5 & 1.6 & 1.1\\
            & GPT4's IK    & 73.2 & 9.7 & 5.6 & 11.4 & 91.3 & 6.0 & 2.2 & 0.6 \\
        \midrule
        \mtr{4}{ar} 
            & Wiki (k=8)   & 52.1 & 17.7 & 4.4 & 25.9 & 72.1 & 20.1 & 1.6 & 6.2 \\
            & Wiki (k=20)  & 55.1 & 14.6 & 5.4 & 24.8 & 75.0 & 17.2 & 1.7 & 6.1 \\
            & Google API   & 60.8 & 8.5 & 11.5 & 19.2 & 81.7 & 8.8 & 6.0 & 3.5 \\
            & GPT4's IK    & 58.9 & 10.9 & 5.5 & 24.7 & 79.6 & 12.6 & 2.8 & 5.0 \\
        \midrule
        \mtr{4}{bn} 
            & Wiki (k=8)   & 43.2 & 38.8 & 0.6 & 17.4 & 52.8 & 44.7 & 0.2 & 2.3 \\
            & Wiki (k=20)  & 51.8 & 30.2 & 0.7 & 17.3 & 63.4 & 34.2 & 0.2 & 2.3 \\
            & Google API   & 72.2 & 9.8 & 3.4 & 14.6 & 86.6 & 10.9 & 1.3 & 1.2 \\
            & GPT4's IK    & 58.1 & 23.9 & 1.3 & 16.7 & 69.3 & 28.2 & 0.7 & 1.8 \\
        \bottomrule
    \end{tabular}
}
\caption{True Positive (TP), False Negative (FN), False Positive (FP), and True Negative (TN) rates for different FActScore pipelines that use GemP as the scorers.}
\label{tab:compare_proposed_method_confusion}
\end{table}

\subsection{Error analysis}

We further conduct an error analysis for the fact-scoring task with these improvements and report in Table \ref{tab:compare_proposed_method_confusion}. The result shows that all three approaches reduce false negatives (and thereby increase true positives) due to their ability to provide more factual coverage. 

Surprisingly, the unverified LLM augmented wikipedia articles significantly increase the true positive rate (by 12.9\%, 6.8\%, and 14.9\% for GemP on es, ar, and bn respectively) without in turn significantly increasing the false positive rate (by 2.6\%, 1.1\% and 0.7\% respectively). In fact, the increase in false positives was lower than using the Google API in all but one case. Conversely, adding additional Wikipedia data always leads to a lower rate of false positives compared to the GPT4 augmented data but also a lower rate of true positives. This implies that the benefits of increased factual coverage from using the unverified GPT4-generated data outweigh the costs of potentially false information introduced. However, these benefits diminish for lower-resource languages, while using the Google API shows more consistent gains across all languages.

\section{Conclusion}
\label{sec:conclusion}

This paper scrutinizes the FActScore pipeline for long-form generated texts in the multilingual setting. We generated new fact candidates and annotated a new corpus for FActScore evaluation. We find that the most recent open-source LLMs struggle with the atomic fact extraction task. Finetuning on this task can match the performance of much larger close-source models, e.g., GPT3.5. More importantly, the Fact Scoring task is very sensitive to the coverage of the knowledge source. Although Wikipedia is reliable, it lacks coverage in lower-resource languages, which leads to a severe underestimation of the FActScore. We show that mitigations such as extending the knowledge source through increasing the amount of Wikipedia data, allowing access to the Internet, and even augmenting low-coverage Wikipedia articles with unverified text generated by an LLM improve multilingual FActScore estimation.

\newpage

\section*{Limitation}

Even though this paper offers insights into the multilingual FActScore, the paper was not able to address more languages than the 3 examined languages and on a larger sample size due to funding limits and the extremely high cost of this task as reported in previous work \cite{min-etal-2023-FActScore,wei2024long}. As a result, the data might contain cultural biases and variations in information and knowledge exposure. Therefore, generalizing our findings to languages other than the examined ones should be considered carefully.
Due to the rapid development of LLMs when the study was done, some models might be obsolete by the publication time, however, we believe this paper still provides insightful knowledge into multilingual factuality scoring.

\section*{Ethical Consideration}

In this work, we hire 6 international crowd-sourced workers from 3 countries as native annotators. The annotators were paid between US\$15 to US\$25 per hour, adjusted to their geographical location. 

While the biographies generated by the two subject models exhibit a certain level of factuality, we observed a significant amount of false information. Using these biographies as references or in real-world scenarios carries the risk of spreading misinformation and negatively impacting the individuals whose biographies are studied.

All the systems presented in this paper do not offer a perfect factual guarantee, especially with the texts and knowledge beyond the studied scope. These systems should be used as alternate tools for traditional factual verification tools.

Given the nature of this task which involves assessing human biographies generated by LLMs, our collected data includes identifications, information, and opinions about them, including false and biased content. We only share the generated texts upon request to enhance the proper use of the data and minimize the risk of spreading false information.

\bibliography{custom,anthology.2020.2022,anthology.2022.2024,anthology.2024.2026}
\bibliographystyle{acl_natbib}

\appendix

\newpage

\section{Biography Selection}
\label{app:biography_geo}

We select a set of people names from the following regions: {\it North America}, {\it Europe}, {\it Asia}, {\it Oceania}, {\it South America}, and {\it Africa};  and 5 levels of rarity based on their Wikipedia page views very frequent, frequent, medium, rare, and very rare.

In Section \ref{sec:knowledge_source}, four additional categories are introduced: {\it internationally popular}, {\it internationally unpopular}, {\it locally popular}, and {\it locally unpopular}. 
The terms {\it locally} and {\it internationally} refer to the geographical or linguistic exposure of the entities whose biographies are being factuality evaluated. 
Local entities might be native speakers of the language or reside in nearby regions where the language is predominantly spoken as a first language. 
For example, for Spanish, this includes regions such as South America and Spain. 
For Arabic, this includes the Arab world, and for Bengali, the Indic region. 
Entities deemed {\it popular} include those classified as {\it very frequent}, {\it frequent} or {\it medium} while {\it unpopular} encompasses {\it medium}, {\it rare}, and {\it very rare} entities according to rarity as introduced above according to Wikipedia page views.

\section{Pilot Experiments on Fact Extractor}
\label{app:fact_extraction_pilot}

We randomly selected 10 sentences from the original work \cite{min-etal-2023-FActScore} and then translated them into target languages. Tested models were prompted (few-shot) to break down those sentences into individual facts. These were translated back to English for assessment based on metrics from Section \ref{sec:fact_extraction}.

Tables \ref{tab:fact_extraction_gpt4}, \ref{tab:fact_extraction_gemini}, \ref{tab:fact_extraction_gpt3.5}, \ref{tab:fact_extraction_mistral}, \ref{tab:fact_extraction_llama2}, \ref{tab:fact_extraction_gemma_instr} represents extractions of GPT4, GemP, GPT3.5, Mistral-7B-Instruct (Mistral), Llama-7B-Chat (Llama2) and Gemma-7B-Instruct respectively. 
All closed models are decent at the task across all studied languages. 
Among open models, Mistral, Llama2, and Gemma could understand the instruction and perform fact extraction, whereas Aya and BLOOMZ were lost in this task (Aya simply returns the original sentence, whereas BLOOMZ does not produce any outputs).
However, in non-English languages, Llama2 shows errors even in a high-resource language like Spanish, while Gemma and Mistral begin to show errors in medium- and low-resource languages. 

For native annotations with R2, we chose two closed models, GPT3.5, and GPT4, and finetuned an open-source model for the extraction task in 3 studied languages. Gemma-7B is chosen considering its large vocab size, thus saving inference costs in the multilingual context. Table \ref{tab:fact_extraction_ft_gemma} illustrates that the finetuned model consistently shows proper extractions across studied languages.
% thanks to its balance in understanding the task and a stronger multilingual LLM.

\section{Open-Source Models Performance as Scorers on More Languages}

\begin{figure}[!h]
    \centering
    \includegraphics[width=\linewidth]{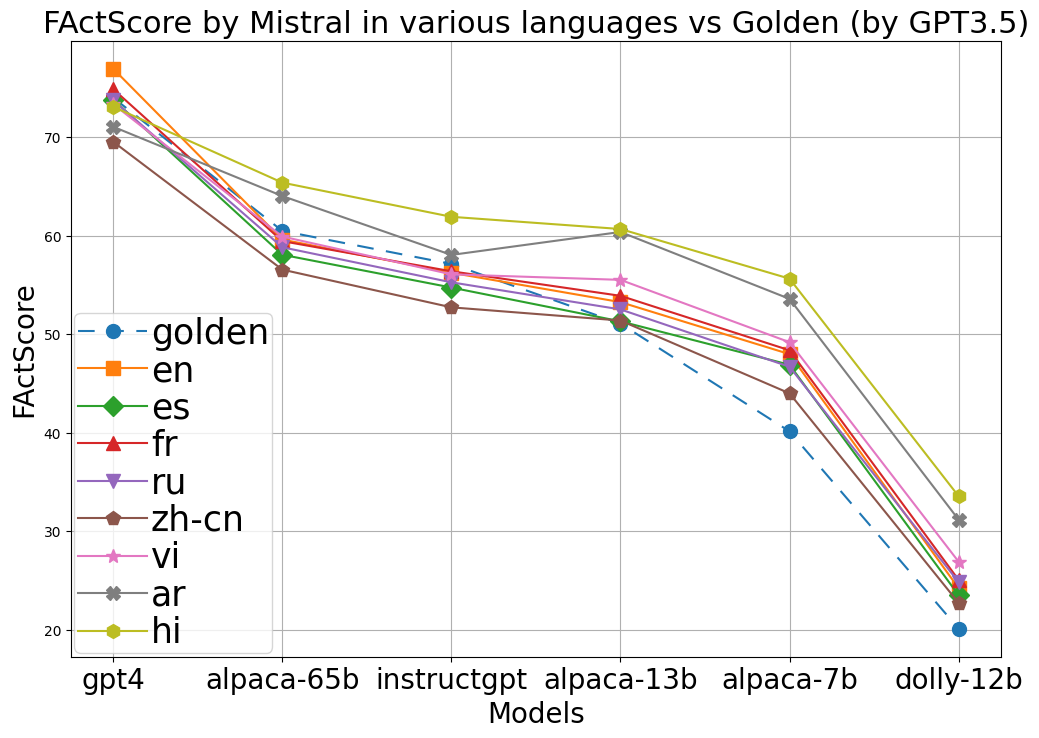} 
    \centering
    \includegraphics[width=\linewidth]{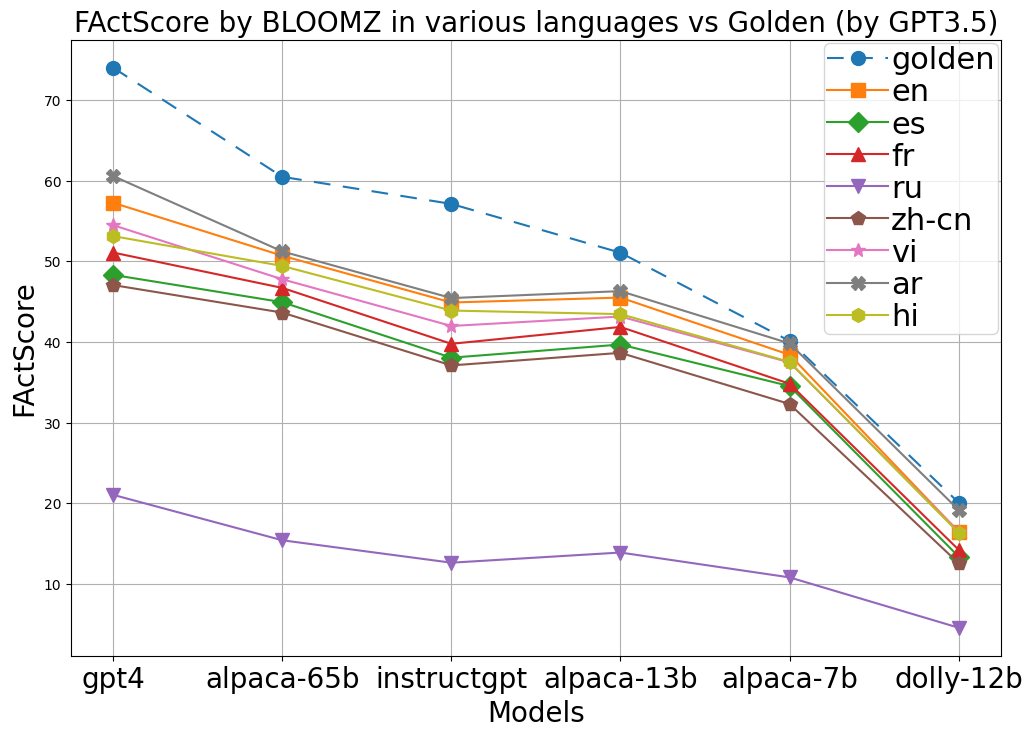} 
    \caption{FActScore by Mistral and BLOOMZ on translated facts generated by studied subject models from \cite{min-etal-2023-FActScore} (R1), compared to golden scoring by GPT3.5, as suggested by \citet{min-etal-2023-FActScore}.}
    \label{fig:FActScore_on_r1}
\end{figure}

\begin{figure}[!h]
    \centering
    \includegraphics[width=\linewidth]{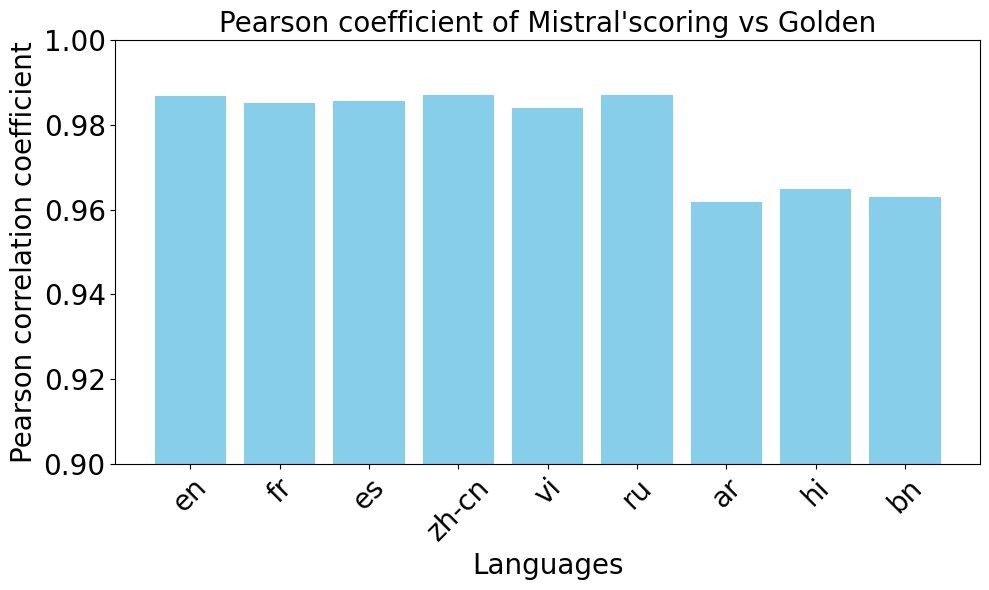} 
    \centering
    \includegraphics[width=\linewidth]{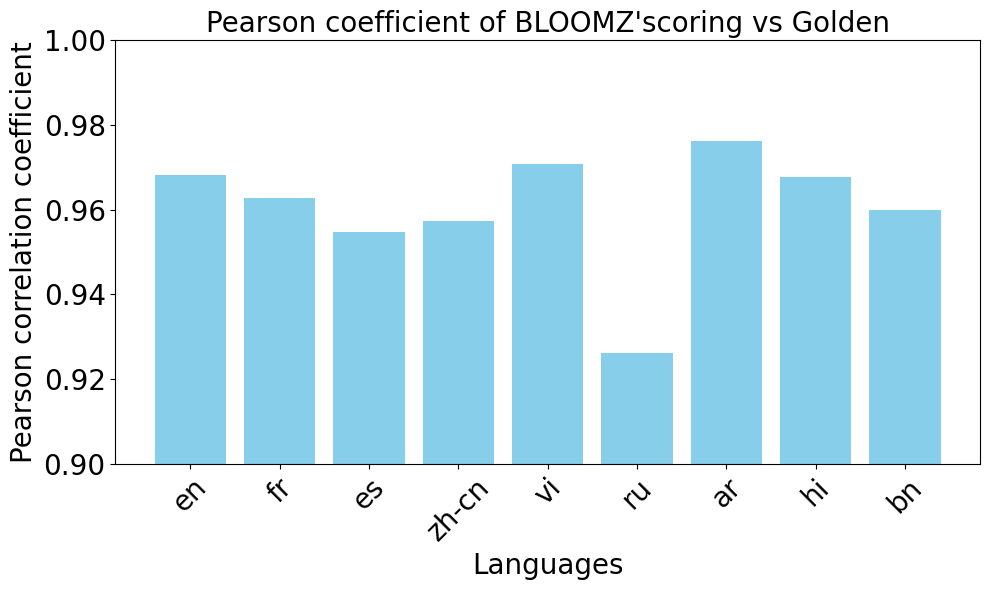} 
    \caption{Pearson correlation coefficient between Mistral (up) and BLOOMZ (down) scoring on subject models from \cite{min-etal-2023-FActScore} with that by GPT3.5 (golden labeling proposed by \cite{min-etal-2023-FActScore}).}
    \label{fig:pearson_vs_golden}
\end{figure}

Figure \ref{fig:FActScore_on_r1} depicts the scoring of subject models by two open-source models, Mistral-7B-Instr (Mistral) and BLOOMZ-7b1 (BLOOMZ) on subject models from \citet{min-etal-2023-FActScore}. 
Both models demonstrate significant agreement in the ranking of subject models when compared to the golden labels provided in the original study \cite{min-etal-2023-FActScore}. It is important to note that the ranking order among evaluated models is the primary concern of \citet{min-etal-2023-FActScore}. 
This is further supported by Figure \ref{fig:pearson_vs_golden}, representing relatively high Pearson correlation coefficients of scoring by two scorers in different languages with golden labeling.

However, there are notable variations in FActScore across languages. 
This indicates that while the pipeline effectively operates in multilingual environments for comparing factuality alignment among language models in a particular language, it is not suitable for assessing model performances across different languages.

\begin{figure}[!h]
    \centering
    \includegraphics[width=\linewidth]{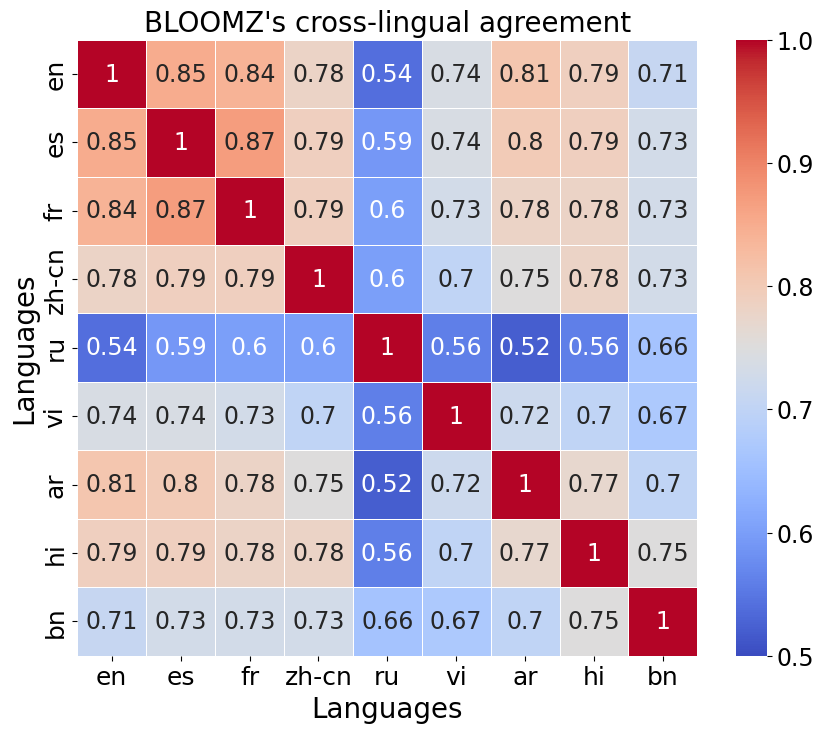}
    \caption{Cross-lingual agreement of Mistral (up) and BLOOMZ (down) when scoring different language versions of the same fact.}
    \label{fig:cross_lingual_agreement_ext}
\end{figure}

Figure~\ref{fig:cross_lingual_agreement_ext} displays the cross-lingual agreement heatmap between texts written in two languages of two open-source models, i.e., Mistral-7B-Instr (Mistral) and BLOOMZ-7b1 (BLOOMZ).
The first row of the heat map illustrates the labeling agreement of both models when evaluating facts in English and non-English languages. 
The agreement for both models decreases in correlation with the resource levels of the non-English languages. 
This decline is clearly observable in Mistral's heat map, but only partially in BLOOMZ's heat map. 
Specifically, BLOOMZ's agreement in Russian and Vietnamese is consistently lower than expected, given their high-resource status in the Common Crawl corpus. 
This issue is attributed to BLOOMZ's alignment training dataset, namely xP3. The xP3 dataset does not include any Russian data and contains a limited amount of Vietnamese data (2.11\% in xP3), less than that for Arabic (2.72\% in xP3), a lower-resource language.

\begin{figure*}[!h]
    \begin{subfigure}{0.24\linewidth}
        \centering
        \includegraphics[width=\linewidth]{figures/cross_lingual_agreement_bloomz_on_translated_facts.png} 
        % \label{fig:cross_lingual_agreement_bloomz}
    \end{subfigure}
    \begin{subfigure}{0.24\linewidth}
        \centering
        \includegraphics[width=\linewidth]{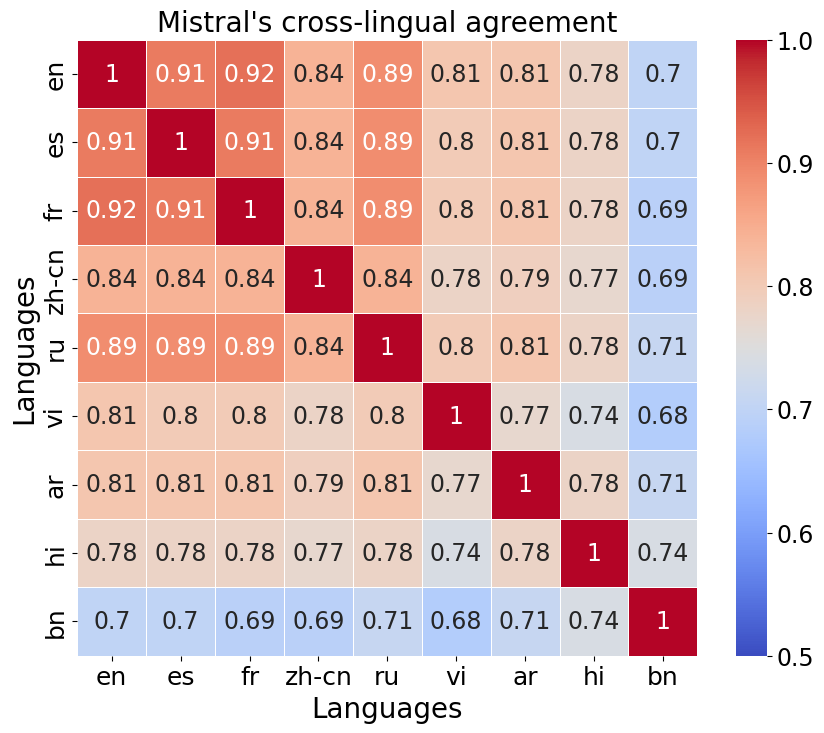} 
        % \label{fig:cross_lingual_agreement_mistral}
    \end{subfigure}
    \begin{subfigure}{0.24\linewidth}
        \centering
        \includegraphics[width=\linewidth]{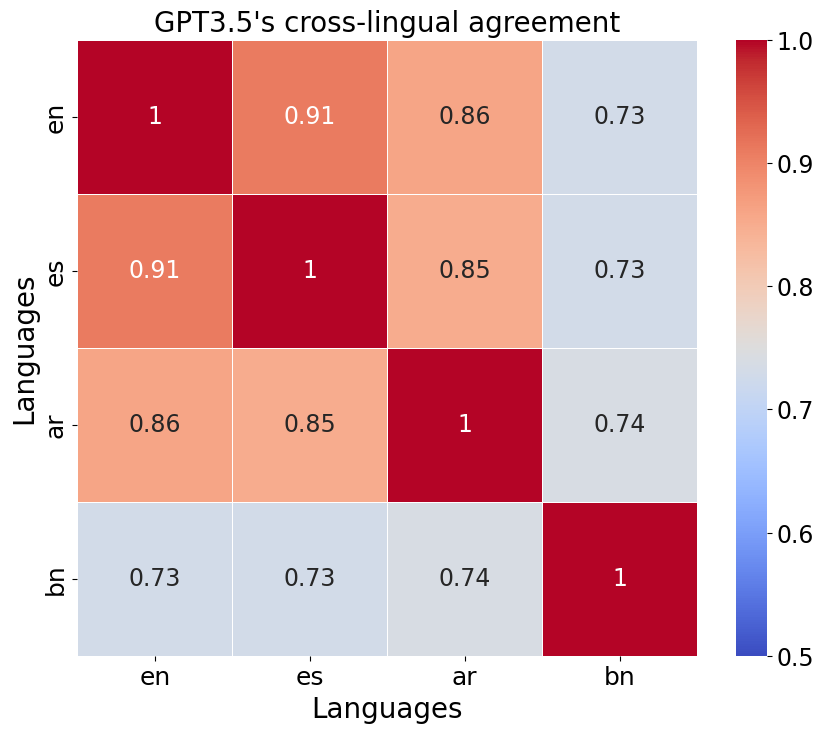}
        % \label{fig:cross_lingual_agreement_gpt_3.5}
    \end{subfigure}
    \begin{subfigure}{0.24\linewidth}
        \centering
        \includegraphics[width=\linewidth]{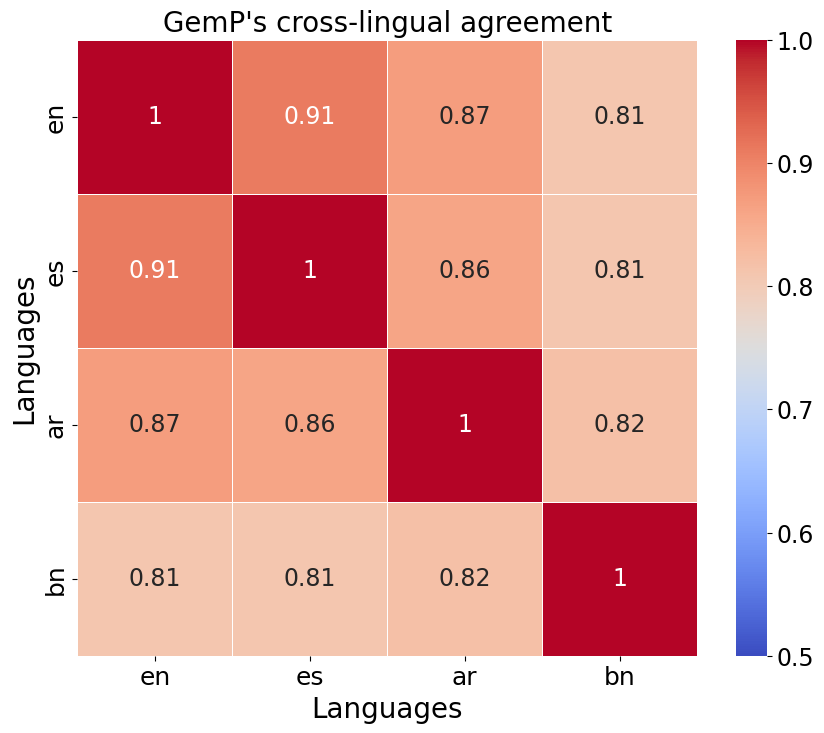} 
        % \label{fig:cross_lingual_agreement_gemini}
    \end{subfigure}
    \caption{Cross-lingual agreement of BLOOMZ (a), Mistral (b), GPT3.5 (c), and GemP (d) when evaluating different language versions of the same fact.}
    \label{fig:cross_lingual_agreement}
\end{figure*}

Figure \ref{fig:cross_lingual_agreement} further illustrates the cross-lingual agreement of two proprietary models, GemP and GPT3.5, with a subset of three out of the nine studied languages. 
The leading open-source model, Mistral, slightly trails behind GPT-3.5, with average scores of 0.83 and 0.85 respectively. 
However, Mistral's performance is significantly lower than that of GemP, which achieves an average score of 0.88.

\section{Impact of Translation on Retriever}
\label{app:impact_translation_retriever}
\begin{table}[!h]
\resizebox{\linewidth}{!}{
\begin{tabular}{llcccc}
        \toprule
        & \mtr{2.2}{\bf Method} & \multicolumn{2}{c}{\bf FActScore} & \multicolumn{2}{c}{\bf Accuracy}\\
        \cmidrule(r){3-4} \cmidrule(r){5-6}
        & & \bf GemP & \bf GPT4 & \bf GemP & \bf GPT4\\
        \midrule
        \mtr{3}{es} & GemP & 58.81 & 68.39 & 72.72 & 73.85\\
                    & GemP (T->R) & 59.97 & 72.92 & 72.18 & 77.66\\
                    & GemP (R->T) & 59.74 & 71.41 & 72.10 & 76.01\\
        \midrule
        \mtr{3}{ar} & GemP & 56.40 & 73.71 & 80.75 & 78.76\\
                    & GemP (T->R) & 52.22 & 69.42 & 80.25 & 78.33\\
                    & GemP (R->T) & 50.79 & 70.06 & 80.00 & 77.47\\
        \midrule
        \mtr{3}{bn} & GemP & 43.78 & 52.97 & 71.89 & 70.63\\
                    & GemP (T->R) & 50.22 & 62.05 & 79.50 & 70.13\\
                    & GemP (R->T) & 39.53 & 49.34 & 70.28 & 65.02\\
        \bottomrule
    \end{tabular}
}
\caption{FActScore and accuracy of performing translation before and after retrieval regarding regarding two metrics. Golden labels are human annotations with 1 Wikipedia page as the knowledge source.}
\label{tab:effect_of_retriever}
\end{table}

Section \ref{sec:fact_extraction} discusses the impact of translation on scoring accuracy by different scorers with clear positive effects on GPT3.5, Mistral, and GemP. 
However, this phenomenon might be attributed to translation's contribution to addressing the multilingual deficiency of the retriever (illustrated in Section \ref{sec:retriever}) as well. This section explores that hypothesis by comparing the effect of translation if it is performed before (T+R) and after retrieval (R+T). 

As shown in Table \ref{tab:effect_of_retriever}, while the difference is not significant in high- and medium-resource languages, for Bengali, performing translation after retrieval (retrieval is in Bengali) significantly diminishes the benefits of translation. Consequently, using translation even results in lower accuracy compared to not using translation at all.

\section{GPT4's Behaviors as a Scorer}

Concurrent with the discussion in Section \ref{sec:qualitative_analysis}, among context-unfaithful samples, there are also factually incorrect ones, including hallucinations and reading deficiencies. A significant portion (72\%) of these factually incorrect samples contains information not found in the knowledge source, hallucination. 
% This implies that the scorer is falsely labeling facts by hallucinating and referring to false information from its internal knowledge.

This category, similar to the discussed factually correct samples, lacks grounded information within the provided context, highlighting an interesting behavior of GPT-4 as a scorer. The model heavily relies on its internal knowledge during the scoring process.

% This reliance might partially explain the decreasing accuracy of GPT4's scoring in lower-resource languages, as specified in Table \ref{tab:FActScore_on_r2} (lower).
% The increasing disparity between the scorer's internal knowledge and the information available in Wikipedia of the studied languages (which decreases with resource levels, as shown in Table \ref{tab:knowledge_source_statistic}) contributes to this issue. 
% Correspondingly, error analysis in Section \ref{sec:qualitative_analysis} reveals a higher number of context-unfaithful samples in lower-resource languages. 
% This indicates GPT-4's increased tendency to rely on its internal knowledge in more limited-resource circumstances.

This reliance may partially explain the decreasing accuracy of GPT-4's scoring in lower-resource languages, as demonstrated in Figure \ref{fig:FActScore_on_r2} (lower). Specifically, Table \ref{tab:knowledge_source_statistic} shows that the information available in the Wikipedia versions of the studied languages diminishes in correlation with their resource levels. This might result in their growing distances with the GPT4's internal knowledge. Consequently, it contributes to lower accuracy (see Figure \ref{fig:FActScore_on_r2} (lower)) when GPT-4 is the scorer.

Correspondingly, error analysis in Section \ref{sec:qualitative_analysis} reveals a higher number of context-unfaithful samples in lower-resource languages. 
This indicates GPT-4's increased tendency to rely on its internal knowledge in more limited-resource circumstances.

Table \ref{tab:error_analysis} illustrates that GPT4 as a scorer is factually correct in about half of the disparity samples with native annotators. 
However, as shown in Table \ref{tab:retriever_vs_without_retriever_agreement_gt}, excluding the retriever and knowledge source from the pipeline and relying solely on GPT4's internal knowledge leads to a decrease in factually correct evaluations overall. 
This implies that despite their limitations, external knowledge sources are essential for maintaining the reliability of the evaluation process.

\begin{table}[!h]
\centering
\begin{tabular}{lcc}
    \toprule
    \bf Lang & \bf \#Facts& \bf \#Passages\\
    \midrule
    es & 391.6 & 15.5\\
    \midrule
    ar  & 317.5 & 13.0\\
    \midrule
    bn & 277.3 & 12.8\\
    \bottomrule
\end{tabular}
\caption{Average number of facts and passages in a Wikipedia page in three languages. }
\label{tab:knowledge_source_statistic}
\end{table}
% \section{Underestimate GPT4 and Natives disagreement attributed to retrieval errors and tabular data}\

\section{Error Analysis Setup}
\label{sec:further_analysis}
For each language, we collected 60 disagreement samples, proportionally distributed according to false positives and false negatives by these model scorers against golden labels by human.

To categorize disagreement cases, we do the following steps:
\begin{itemize}
    \item Thoroughly read the entire Wikipedia article to identify relevant text (sentences, paragraphs) for evaluating the fact and checking for annotator errors.
    \item If no text within the Wikipedia page relates to the fact, it should be labeled as ``not supported'' by annotators (or it would be a mistake from the annotator) and ``supported'' by the model scorer. We then proceed to evaluate the fact based on external sources and determine whether the labeling should be classified as ``context unfaithful but factually correct'' (if supported by external sources) or ``context unfaithful and hallucinated'' (if not supported by external sources).
    \item If related information is found within the Wikipedia page, classify the labeling disagreements as follows:
    \begin{itemize}
        \item Tabular data: The information is in a table and has not been processed by Wikipedia's HTML conversion to text.
        \item Retriever error: The information is not in the passages retrieved.
        \item The information is in the retrieved passages but missed by scorers.
        \item Cannot Deduct from Context: Correct evaluation of the fact, while not being explicitly specified, but deductible from the provided context, but the modeling evaluator fails to do so.
        \item Subjective opinion: The labeling is hugely influenced by the annotator's subjective opinion.
    \end{itemize}
    \item Other cases to consider:
    \begin{itemize}
        \item Assistant Generation: If the sentence is part of the model's service generated content.
        % \item Task/Annotation Design: If the disparity arises because human annotators consider the whole sentence context, while the model labels individual facts, leading to labeling differences.
    \end{itemize}
\end{itemize}

\begin{table*}[!h]
\resizebox{\linewidth}{!}{
    \begin{tabular}{ll|cc|cc|cc|cc|cc|cc}
        \toprule
        \mtr{4}{\bf Scorer} & \mtr{4}{\bf Reference} & \multicolumn{6}{|c|}{\bf FActScore} & \multicolumn{6}{c}{\bf Accuracy}\\
        \cmidrule(r){3-8} \cmidrule(r){9-14}
        & & \multicolumn{2}{|c|}{\bf es} & \multicolumn{2}{c|}{\bf ar} & \multicolumn{2}{c|}{\bf bn} & \multicolumn{2}{c|}{\bf es} & \multicolumn{2}{c|}{\bf ar} & \multicolumn{2}{c}{\bf bn}\\
        \cmidrule(r){3-4} \cmidrule(r){5-6} \cmidrule(r){7-8} \cmidrule(r){9-10} \cmidrule(r){11-12} \cmidrule(r){13-14}
        & & \bf GemP & \bf GPT4 & \bf GemP & \bf GPT4 & \bf GemP & \bf GPT4 & \bf GemP & \bf GPT4 & \bf GemP & \bf GPT4 & \bf GemP & \bf GPT4\\
        \midrule
        \mtr{3}{Human} & Internet & 82.7 & 97.3 & 70.4 & 92.2 & 81.8 & 97.5 & 100 & 100 & 100 & 100 & 100 & 100\\
        & Wikipedia (1 page) & 75.4 & 88.4 & 61.6 & 81.5 & 57.5 & 60.2 & 92.1 & 91.1 & 91.7 & 91.1 & 89.3 & 90.4\\
        & Wikipedia (All) & 78.3 & 91.7 & 64.4 & 86.5 & 60.1 & 60.6 & 95.0 & 94.2 & 93.7 & 94.3 & 76.4 & 72.5\\
        \midrule
        \mtr{4}{GPT4} & w/ Wikipedia & 78.2 & 91.7 & 68.4 & 93.5 & 74.0 & 91.7 & 86.2 & 91.0 & 85.9 & 93.3 & 86.0 & 90.9\\
        & w/o Wikipedia & 87.6 & 97.1 & 87.3 & 98.2 & 85.4 & 93.7 & 86.0 & 94.7 & 75.3 & 91.5 & 79.9 & 91.6\\
        & T + w/ Wikipedia & 78.2 & 94.4 & 65.7 & 91.6 & 73.4 & 86.3 & 84.9 & 93.7 & 84.1 & 92.0 & 84.0 & 86.4\\
        & T + w/o Wikipedia & 95.9 & 95.3 & 79.3 & 95.5 & 81.1 & 82.9 & 85.0 & 93.3 & 79.0 & 89.9 & 82.8 & 83.1\\
        \midrule
        \mtr{4}{GemP} & w/ Wikipedia & 63.3 & 72.8 & 56.4 & 73.7 & 43.8 & 53.0 & 74.3 & 74.0 & 77.9 & 78.3 & 60.6 & 55.1\\
        & w/o Wikipedia & 79.4 & 84.4 & 74.1 & 88.4 & 77.3 & 86.2 & 76.2 & 82.7 & 68.5 & 82.1 & 72.7 & 84.1\\
        & T + w/ Wikipedia & 60.4 & 73.0 & 53.4 & 69.7 & 49.9 & 61.9 & 69.0 & 73.9 & 76.3 & 74.9 & 67.0 & 63.4\\
        & T + w/o Wikipedia & 75.6 & 80.0 & 67.9 & 84.6 & 71.5 & 73.6 & 73.1 & 78.7 & 68.7 & 80.3 & 72.5 & 73.1\\
        \midrule
        \mtr{4}{GPT3.5} & w/ Wikipedia & 81.7 & 90.9 & 69.2 & 88.9 & 59.1 & 71.1 & 81.3 & 89.8 & 79.5 & 88.9 & 65.2 & 70.0\\
        & w/o Wikipedia & 84.3 & 91.5 & 82.3 & 93.6 & 74.6 & 80.9 & 79.3 & 89.8 & 67.6 & 87.0 & 71.5 & 78.8\\
        & T + w/ Wikipedia & 84.8 & 93.4 & 74.3 & 92.8 & 73.0 & 83.3 & 81.7 & 91.8 & 82.6 & 91.9 & 79.6 & 83.1\\
        & T + w/o Wikipedia & 84.7 & 92.2 & 77.9 & 93.4 & 82.8 & 84.7 & 78.0 & 90.4 & 77.1 & 88.0 & 80.2 & 83.9\\
        \midrule
        \mtr{4}{Mistral} & w/ Wikipedia & 72.1 & 84.4 & 57.5 & 78.5 & 45.0 & 59.5 & 77.9 & 84.0 & 73.4 & 80.6 & 55.6 & 59.0\\
        & w/o Wikipedia & 58.6 & 71.0 & 43.9 & 65.2 & 60.9 & 72.1 & 61.2 & 69.9 & 54.5 & 61.1 & 57.2 & 71.6\\
        & T + w/ Wikipedia & 74.6 & 86.9 & 61.0 & 85.8 & 61.2 & 74.5 & 78.2 & 86.2 & 81.6 & 88.2 & 73.3 & 76.3\\
        & T + w/o Wikipedia & 67.8 & 77.7 & 61.0 & 82.5 & 65.2 & 67.7 & 69.3 & 76.9 & 68.6 & 80.0 & 68.7 & 68.2\\
        \bottomrule
    \end{tabular}
}
\caption{FActScore and accuracy by different scorers with (w/) or without (w/o) Wikipedia and whether translation (T) is used on generated facts and knowledge source (Wikipedia page). Accuracy is measured against natives labeling using the Internet to find references.}
\label{tab:retriever_vs_without_retriever_agreement_gt}
\end{table*}

\begin{table*}[h!]
\small
\centering
\begin{tabular}{lccccccccc}
\toprule
Recall@5 & en & fr & es & ru & zh-cn & vi & ar & hi & bn \\
\midrule
distiluse-base-multilingual-cased-v2 & 67.70 & 66.02 & 66.13 & 66.14 & 64.69 & 65.42 & 63.37 & 61.54 & 51.19 \\
paraphrase-multilingual-MiniLM-L12-v2 & 65.76 & 63.79 & 63.82 & 63.18 & 63.77 & 63.42 & 60.88 & 61.40 & 54.67 \\
\bottomrule
\end{tabular}
\caption{Recall@5 scores by different multilingual versions of Sentence-BERT. Retrieved passages by the original retriever in English \cite{ni2021large} is considered golden to calculate Recall@5}
\label{tab:compare_retriever}
\end{table*}

\section{Experimental Settings}
We utilized and assessed the following models to study components of the FActScore pipeline.

{\bf Subject Models}:
\begin{itemize}
    \item GemP (gemini-1.0-pro)
    \item GPT4 (gpt-4-0125-preview)
\end{itemize}

{\bf Factuality Scorers}:
\begin{itemize}
    \item GemP (gemini-1.0-pro)
    \item GPT4 (gpt-4-0125-preview)
    \item GPT3.5 (gpt-35-turbo-0125)
    \item Mistral (mistralai/Mistral-7B-Instruct-v0.2)
    \item BLOOMZ (bigscience/bloomz-7b1)
\end{itemize}

{\bf Fact Extractors}:
\begin{itemize}
    \item GemP (gemini-1.0-pro)
    \item GPT4 (gpt-4-0125-preview)
    \item GPT3.5 (gpt-35-turbo-0125)
    \item Gemma (google/gemma-7b-it)
    \item Mistral  (mistralai/Mistral-7B-Instruct-v0.2)
    \item Aya (CohereForAI/aya-101)
    \item BLOOMZ (bigscience/bloomz-7b1)
    \item Llama2 (meta-llama/Llama-2-7b-chat-hf)
\end{itemize}

{\bf Retrievers}:
\begin{itemize}
    \item sentence-transformers/paraphrase-multilingual-MiniLM-L12-v2
    \item sentence-transformers/distiluse-base-multilingual-cased-v2
\end{itemize}

{\bf Knowledge Generator}:
\begin{itemize}
    \item GPT4 (gpt-4-0125-preview)
\end{itemize}

{\bf Translator}:
\begin{itemize}
    \item Google Translate (Cloud Translation - Basic (v2), used from January 2024 to June 2024)
\end{itemize}

{\bf Running Trials of Experiments}: All results were obtained from data conducted or collected from a single trial.

\section{Hyper-Parameters}

All experiments are conducted from January to June 2024. The following hyper-parameters are specified, while all others are set to their default values.

{\bf Generation Temperature}: All studied models' temperatures are set to 0.7.

{\bf Context, max generation length}: For open-source models, the maximum output length is set to 512 tokens, and the maximum sequence length is set to 4096 tokens for high-resource languages, and 1024 and 6024 tokens for medium- and low-resource languages, respectively. For closed models accessed via API, the maximum token limit is uniformly set to 4096 tokens for all use cases.
\section{Instructions for Data Annotation}
\subsection{Factuality Labeling}
To collect R2, the original pipeline from \citet{min-etal-2023-FActScore} is fully replicated to studied languages. Along with it, we had the qualification task to assess annotators and provided an 1-hour training session.
\subsection{Fact Extraction}
\begin{figure*}[t]
    \begin{subfigure}{\linewidth}
        \includegraphics[width=\linewidth]{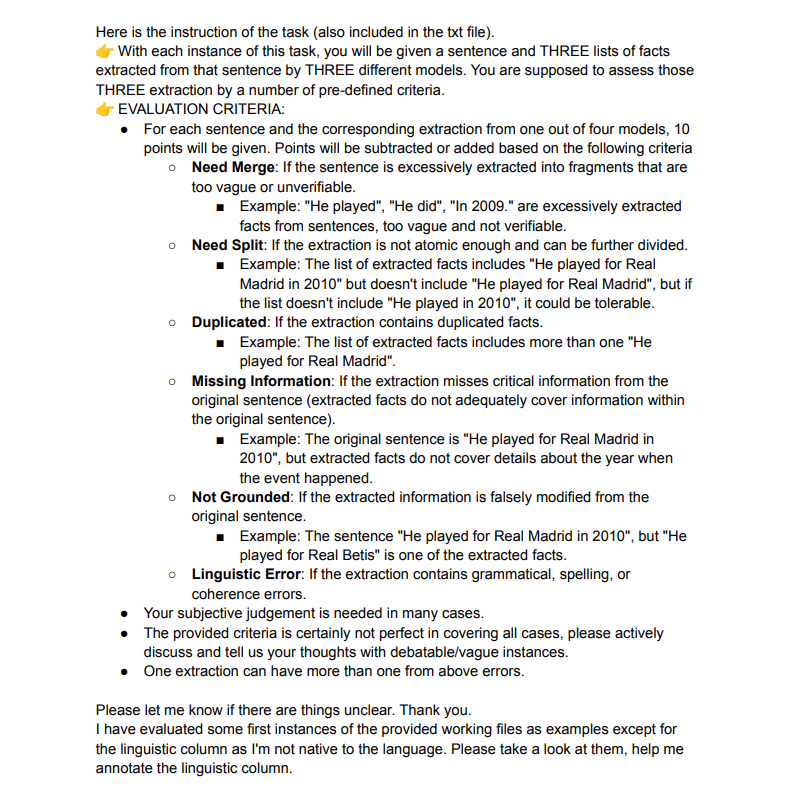}
    \end{subfigure}
    \caption{Instructions for data annotation in Section \ref{sec:fact_extraction} on Fact Extraction component.}
    \label{fig:rule_extraction}
\end{figure*}

In the additional task on the fact extraction component, discussed in Section \ref{sec:fact_extraction}, native annotators followed the guideline outlined in Figure \ref{fig:rule_extraction}.
\clearpage
\begin{table*}[!h]
    \small
    \centering
    \begin{tabular}{p{0.1\linewidth}p{0.2\linewidth}p{0.6\linewidth}}
    \toprule
        \bf Category & \bf Candidate Fact & \bf Comment\\
        \midrule
        \bf Context unfaithful - Factually correct 
        & {\it``Naipaul fue criticado por su visión a menudo pesimista''}
        
        {\bf Translated:} {\it``Naipaul was criticized for his often pessimistic vision''}
        & Native label: F, Model label: T, Ground truth: T.
        
        {\bf Comment:} No related information within the provided Wikipedia page. But there is supporting evidence from en.wikipedia.
        
        {\bf Evidence:} {\it``Yet he has been accused of being a``neo-colonialist'' , and in this novel post-colonial Africa is depicted as spiraling into a kind of Hell...Naipaul's fiction and especially his travel writing have been criticised for their allegedly unsympathetic portrayal of the Third World. The novelist Robert Harris has called Naipaul's portrayal of Africa racist and``repulsive,''reminiscent of Oswald Mosley's fascism.''} \\
        \midrule
        \bf Context unfaithful - Reading Deficiency & \it{``Rodrygo Goes de Souza nació el 9 de enero de 2001.''}
        
        {\bf Translated:}  {\it``Rodrygo Goes de Souza was born on January 9, 2001.''} & Native label: T, Model label: F, Ground truth: T.
        
        {\bf Comment:} The evaluator misses the related information (that supports the fact) within retrieved passages.
        
        {\bf Evidence:} {\it``Rodrygo Silva de Goes (; Osasco, São Paulo, 9 de enero de 2001), conocido simplemente como Rodrygo, es un futbolista brasileño que juega como delantero en el Real Madrid C. F. de la Primera División de España.''}
        
        {\bf Translated:} {\it``Rodrygo Silva de Goes (; Osasco, São Paulo, January 9, 2001), known simply as Rodrygo, is a Brazilian footballer who plays as a forward for Real Madrid C.F. of the Spanish First Division.''} \\
        \midrule
        \bf Retrieval error & {\it``Ingresó al seminario de Villa Devoto en Buenos Aires.''}
        
        {\bf Translated:} {\it``He entered the Villa Devoto seminar in Buenos Aires.''}
        & Native label: T, Model label: F, Ground truth: T.
        
        {\bf Comment:} The retriever fails to retrieve the needed information passage for evaluation.
        
        {\bf Evidence:} {\it``Ingresó al seminario del barrio Villa Devoto y al noviciado de la Compañía de Jesús.''}
        
        {\bf Translated:} {\it``He entered the seminary in the Villa Devoto neighborhood and the novitiate of the Society of Jesus.''} \\
        \midrule
        \bf Tabular data & {\it``El primer torneo importante que Court ganó fue el campeonato australiano de tenis.''}

        {\bf Translated:} {\it ``The first major tournament Court won was the Australian Tennis Championships.''}  & Native label: T, Model label: F, Ground truth: T.
        
        {\bf Comment:} Related information is embedded in the table which is not processed by Wikipedia's HTML conversion to text, thus not being contained as passages to retrieve.
        
        {\bf Evidence:} Information lies within``Victorias (24)''table at the first row\\
        \midrule
        \bf Subjective opinion & {\it ``Es considerado uno de los trompetistas más destacados de su generación.''}
        
        {\bf Translated:} {\it ``He is considered one of the most prominent trumpeters of his generation.''}  
        & Native label: T, Model label: F, Ground truth: T/F.
        
        {\bf Comment:} The statement/fact is subjective, thus debatable. Chuck Mangione had a song, being recognized as the number one jazz song of all time by a radio channel, but there is no explicit mention that he is a prominent trumpeter of his generation.
        
        {\bf Evidence:} {\it``Recientemente las estaciones de radio que transmiten jazz en los Estados Unidos han reconocido a Feels So Good de Mangione como la canción número uno de todos los tiempos.''}({\bf en:} {\it``Recently, jazz radio stations in the United States have recognized Mangione's Feels So Good as the number one song of all time.''} \\
        \midrule
        \bf Annotation error & {\it``Tekke dio el salto al fútbol europeo en 2006.''}
        
        {\bf Translated:}  {\it``Tekke made the leap to European football in 2006.''} 
        & Native label: F, Model label: T, Ground truth: T.
        
        {\bf Comment:} The annotator misses details within the Wikipedia page.
        
        {\bf Evidence:} {\it``Esa temporada Tekke se convirtió en el máximo goleador de la Superliga de Turquía al anotar 31 goles. El 31 de julio de 2006 firma un contrato con su actual club, el Zenit de San Petersburgo ruso, equipo que realizó un desembolso económico de 10 millones de euros para poder hacerse con sus servicios.''}
        
        {\bf Translated:} {\it``That season Tekke became the top scorer in the Turkish Super League by scoring 31 goals. On July 31, 2006, he signed a contract with his current club, the Russian Zenit Saint Petersburg, a team that made a financial outlay of 10 million euros to be able to acquire his services.''}\\
        \midrule
        \bf Assistant generation & {\it``La información podría haberse modificado.''}
        
        {\bf Translated:} {\it``The information may have been modified.''}
        & Native label: F, Model label: T, Ground truth: F.
        
        {\bf Comment:} The assistant/service generation by subject models is often labeled as {\it ``supported''} by modeling evaluators.\\
        \bottomrule
    \end{tabular}
\caption{Examples from each disagreement category between natives and Gemini in Spanish.}
\label{tab:example_disgreement_gemini_es}
\end{table*}

\begin{table*}[!h]
    \small
    \centering
    \begin{tabular}{p{0.1\linewidth}p{0.2\linewidth}p{0.65\linewidth}}
        \toprule
        \bf Category & \bf Fact & \bf Comment\\
        \midrule
        \bf Context unfaithful - Factually correct & \langar{ هولك هوجان} 
        
        {\bf Translated:} {\it ``He had a big mustache''}
        & Native label: F, Model label: T, Ground truth: T.
        
        No related information within the provided Wikipedia page. But there is supporting evidence from en.wikipedia.
        
        {\bf Evidence:} {\it ``Hogan grew a beard alongside his famous mustache and dyed it black, traded his red and yellow garb in for black and white clothing, often detailed with lightning bolts, and renamed himself "Hollywood" Hulk Hogan (often shortened to Hollywood Hogan''}\\
        \midrule
        \bf Context unfaithful - Hallucination & \langar{الرابطة الوطنية لحقوق المرأة. }
        
        {\bf Translated:} {\it ``Stone helped found the National Women's Suffrage Association.''}
        & Native label: F, Model label: T, Ground truth: F.
        
         The fact is false. Lucy Stone helped found the American Woman Suffrage Association (AWSA) in 1869, which was a rival organization to the National Woman Suffrage Association (NWSA) founded by Elizabeth Cady Stanton and Susan B. Anthony\\
        \midrule
        \bf Context unfaithful - Reading Deficiency & \langar{الاهتمامات البحثية تشمل نظرية ما بعد الاستعمار} 
        
        {\bf Translated:} {\it ``Research interests include postcolonial theory.''}
        & Native label: T, Model label: F, Ground truth: T.
        
        {\bf Comment:} The evaluator misses the related information (that supports the fact) within retrieved passages.
        
        {\bf Evidence:} \langar{المنشور عام 1985 يعد من النصوص المؤسسة لـ ما بعد الكولونيالية، وتعد سبيفاك حاليًا من أهم الشخصيات العالمية المؤثرة في النقد الحضاري والأدب. حصلت سبيفاك على جائزة كيوتو للفنون والفلسفة لعام 2012 لكونها «عالمة نظريات ناقدة (منظرة ناقدة) ومعلمة تدافع عن العلوم الإنسانية ضد الاستعمار الفكري فيما يتعلق بالعالم المعولم }
        
        {\bf Translated:}  {\it ``The 1985 publication is considered one of the founding texts of postcolonialism, and Spivak is currently considered one of the most important international figures influencing cultural criticism and literature. Spivak was awarded the 2012 Kyoto Prize for Arts and Philosophy for being a "critical theorist (critical theorist) and educator who defends the humanities against intellectual colonialism in relation to the globalized world.}\\
        
        \midrule
        \bf Retrieval error & \langar{كانت صفقة انتقال نيمار بقيمة 222 مليون يورو.}" 
        
        {\bf Translated:} {\it ``Neymar's transfer was worth 222 million euros.''}
        & Native label: T, Model label: F, Ground truth: T.
        
        {\bf Comment:} The retriever fails to retrieve needed information passage for evaluation.
        
        {\bf Evidence:} \langar{في عام 2017، انتقل نيمار إلى باريس سان جيرمان في صفقة قياسية هي الأضخم في تاريخ كرة القدم، حيث بلغت قيمتها 222 مليون يورو} ({\bf en:} "In 2017, Neymar moved to Paris Saint-Germain in a record deal, the largest in football history, worth 222 million euros.")\\
        \midrule
        \bf Tabular data & \langar{متزوج من الملكة سوثيدا فاجيرالونجكورن نا أيوديا.} 
        
        {\bf Translated:} {\it ``Married to Queen Suthida Vajiralongkorn na Ayodhya.''} & Native label: T, Model label: F, Ground truth: T.
        
        {\bf Comment:} Related information is embedded in the table which is not processed by Wikipedia's HTML conversion to text, thus not being contained as passages to retrieve.
        
        {\bf Evidence:} Information locates on \langar{الزوجة} (Wife) section of the side infobox.\\
        \midrule
        \bf Subjective opinion & \langar{هند صبري تعتبر مثالًا يحتذى به.} 
        
        {\bf Translated:} {\it ``Hend Sabry is a role model.''}
        & Native label: T, Model label: F, Ground truth: T/F.
        
        {\bf Comment:} The statement/fact is subjective, thus debatable.\\
        \midrule
        \bf Assistant generation & \langar{لم تتوفر معلومات محددة حول وفاة عبد القادر الشاوي حتى عام 2023.} 
        
        {\bf Translated:} {\it ``No specific information about Abdelkader Chaoui's death is available until 2023.''}
        & Native label: F, Model label: T, Ground truth: F.
        
        {\bf Comment:} The assistant/service generation by subject models is often labeled as ``supported'' by modeling evaluators.\\
        \midrule
        % \bf Annotation error & "كان عمل تيدروس لتعزيز الشفافية." 
        
        % {\bf Translated:} {\it ``Tedros' work was to promote transparency.''} & Native label: T, Model label: F, Ground truth: F.
        
        % {\bf Comment:} There is no evidences that Tedros' work was to promote transparency, he was even being criticized due to the lack of transparency
        
        % {\bf Evidence:} "بينما تلقى تيدروس المديح لالتزامه بالمساواة بين الجنسين، تلقى أيضًا انتقادات لافتقاره إلى الشفافية. إذ عيَّن الدكتورة تريزا كاسييفا من وزارة الصحة الروسية لقيادة برنامج منظمة الصحة العالمية لمكافحة السل دون التماس مساهمات المجتمع المدني؛ قبل أيام من التعيين، نشرت منظمات المجتمع المدني رسالة مفتوحة تدعو إلى عملية تنافسية مفتوحة لتحديد المدير الجديد للبرنامج."
        
        % {\bf Tranlated:} {\it ``While Tedros has received praise for his commitment to gender equality, he has also received criticism for his lack of transparency. He appointed Dr. Teresa Kasyeva from the Russian Ministry of Health to lead the WHO tuberculosis program without soliciting civil society contributions; Days before the appointment, civil society organizations published an open letter calling for an open, competitive process to identify the new program director.''}\\
        % \midrule
        \bf Inconsistent Wikipage & \langar{جائزة الكرة الذهبية فاز بها 7 مرات.}
        
        {\bf Translated:} {\it ``The Ballon d'Or he won 7 times.''}
        & Native label: F, Model label: T, Ground truth: F.
        
        {\bf Comment:} The Wikipedia page has conflicting information on the number of Ballon d'Or that Lionel Messi won.\\
        \bottomrule
    \end{tabular}
\caption{Examples from each disagreement category between natives and Gemini in Arabic.}
\label{tab:example_disgreement_gemini_ar}
\end{table*}

\begin{table*}[!h]
\small
    \centering
    \begin{tabular}{p{0.1\linewidth}p{0.2\linewidth}p{0.65\linewidth}}
    \toprule
        \bf Category & \bf Fact & \bf Comment\\
        \midrule
        \bf Context unfaithful - Factually correct & {\it Trygve Lie intentó promover la paz.''}
        
        {\bf Translated:} {\it ``Trygve Lie tried to promote peace.''} 
        & Native label: F, Model label: T, Ground truth: T.
        
        No related information within the provided Wikipedia page. But there is supporting evidence from en.wikipedia.
        
        {\bf Evidence:}  {\it ``He sent 50 members of the United Nations guard force from Lake Success to assist the Mediator in supervising the Truce in the former British Mandate of Palestine in 1948 and the "UNTSO", the first peacekeeping operation was established by the United Nations.''}\\
        \midrule
        \bf Context unfaithful - Hallucination & {\it Algunas de sus películas han recibido críticas positivas a nivel internacional.''}
        
        {\bf Translated:} {\it ``Some of his films have received positive reviews internationally.''} 
        & Native label: F, Model label: T, Ground truth: F.
        
         The fact is false. Besides several users on IMDB and Rotten Tomatoes, there is no concrete evidence that supports Surya Saputra’s films are recognized internationally.\\
        \midrule
        \bf Context unfaithful - Reading Deficiency & {\it ``Murió de un ataque al corazón.''}
        
        {\bf Translated:} {\it ``He died of a heart attack.''}
        
        & Native label: F, Model label: T, Ground truth: F.
        
        {\bf Comment:} He died of pneumonia, not a heart attack.
        
        {\bf Evidence:} {\it ``...Arruinado, físicamente débil y con la mente deteriorada, Capone se retiró a una propiedad ubicada en Palm Island, en Miami Beach, Florida, donde se recluyó con su esposa del mundo exterior. El 21 de enero de 1947, sufrió un derrame cerebral, y murió cuatro días después de neumonía: Al Capone fue encontrado muerto en la bañera''}
        
        ({\bf en:} {\it ``...Ruined, physically weak and mentally deteriorating, Capone retired to a property located on Palm Island in Miami Beach, Florida, where he and his wife secluded themselves from the outside world. On January 21, 1947, he suffered a stroke, and died four days later of pneumonia: Al Capone was found dead in the bathtub.}\\
        \midrule
        \bf Retrieval error & {\it ``Incluyó su papel en \"Guardianes de la Galaxia Vol. 2014''}
        
        {\bf Translated:} {\it ``Included his role in \"Guardians of the Galaxy Vol. 2014.''''} & Native label: T, Model label: F, Ground truth: T.
        
        {\bf Comment:} The retriever fails to retrieve the needed information passage for evaluation.
        
        {\bf Evidence:} {\it ``En 2014, logró el reconocimiento a nivel mundial al protagonizar la película Guardianes de la Galaxia (2014) con el papel de Peter Quill / Star-Lord.23 El filme recibió elogios de la crítica por su humor y fue un éxito comercial tras recaudar 773 millones de dólares, además de convertirse en la cuarta película más taquillera de 2014''}
        
        {\bf Translated:} {\it ``In 2014, he achieved worldwide recognition by starring in the film Guardians of the Galaxy (2014) with the role of Peter Quill / Star-Lord.23 The film received critical praise for its humor and was a commercial success after grossing \$773. million dollars, in addition to becoming the fourth highest-grossing film of 2014''}\\
        \midrule
        
        \bf Tabular data & {\it ``Drummond promedió 17.5 puntos por partido en la temporada 2020-21.''}
        
        {\bf Translated:} {\it ``Drummond averaged 17.5 points per game in the 2020-21 season.''} 
        
        & Native label: T, Model label: F, Ground truth: T.
        
        {\bf Comment:} Related information is embedded in the table which is not processed by Wikipedia's HTML conversion to text, thus not being contained as passages to retrieve.\\
        \midrule
        \bf Subjective opinion & {\it ``Sarr es hábil.''}
        
        {\bf Translated:} {\it ``Sarr is skillful''} 
        
        & Native label: F, Model label: T, Ground truth: T/F.
        
        {\bf Comment:} The statement/fact is subjective, thus debatable.\\
        \midrule
        \bf Assistant generation & {\it ``Nuevos proyectos y logros pueden haberse agregado a la biografía de Surya Saputra después de 2023.''}
        
        {\bf Translated:} {\it ``New projects and achievements may have been added to Surya Saputra's biography after 2023.''} 
        & Native label: F, Model label: T, Ground truth: F.
        
        {\bf Comment:} The assistant/service generation by subject models is often labeled as ``supported'' by modeling evaluators.\\
        % \midrule
        % \bf Annotation error & {\it ``Tekke destacó por su habilidad para asistir a sus compañeros.}
        
        % {\bf Translated:} {\it ``Tekke stood out for his ability to assist his teammates.''} 
        % & Native label: T, Model label: F, Ground truth: F.
        
        % {\bf Comment:} There is no evidence on Fatih Tekke’s capability to assist his teammate.\\
        \bottomrule
    \end{tabular}
\caption{Examples from each disagreement category between natives and GPT-4 in Spanish.}
\label{tab:example_disgreement_gpt4_es}
\end{table*}

\begin{table*}[!h]
    \small
    \centering
    \begin{tabular}{p{0.1\linewidth}p{0.2\linewidth}p{0.65\linewidth}}
    \toprule
        \bf Category & \bf Fact & \bf Comment\\
        \midrule
        \bf Context unfaithful - Factually correct & \langar{والده كان عازفًا.}
        
        {\bf Translated:} {\it ``His father was a musician.''} & Native label: F, Model label: T, Ground truth: T.
        
        No related information within the provided Wikipedia page. But there is supporting evidence from en.wikipedia.
        
        Evidence: His mother is dancer Kine Gueye Thiam (née Gueye), and his father is percussionist Mor Thiam. Mor Thiam was born to a Toucouleur family of Quranic scholars in Kaolack, Senegal.\\
        \midrule
        \bf Context unfaithful - Hallucination & \langar{هو الابن الأكبر.}
        
        {\bf Translated:} {\it ``He is the eldest son.''} 
        & Native label: F, Model label: T, Ground truth: F.
        
         The fact is false. The eldest son of his father is Abdelaziz bin Khalifa Al Thani\\
        \midrule
        \bf Context unfaithful - Reading Deficiency & \langar{توفيت في 8 سبتمبر 2022.}
        
        {\bf Translated:} {\it ``She died on September 8, 2022.''} & Native label: T, Model label: F, Ground truth: T.
        
        {\bf Comment:} The evaluator misses the related information (that supports the fact) within retrieved passages.
        
        {\bf Evidence:} \langar{في 8 سبتمبر 2022، أعلن قصر باكنغهام وفاة الملكة إليزابيث الثانية عن عمر يناهز 96 عاما، تزامن ذلك مع أنباء متواترة حول تدهور حالتها الصحية.}
        
        ({\bf Translated:} {\it ``On September 8, 2022, Buckingham Palace announced the death of Queen Elizabeth II at the age of 96, coinciding with frequent reports about the deterioration of her health.''} \\
        \midrule
        \bf Retrieval error & \langar{اسم الطفل الثاني هو روري جون جيتس.}
        
        {\bf Translated:} {\it ``The name of the second child is Rory John Gates.''} & Native label: T, Model label: F, Ground truth: T.
        
        {\bf Comment:} The retriever fails to retrieve needed information passage for evaluation.
        
        {\bf Evidence:} \langar{تزوج بيل غيتس من ميليندا فرينش في عام 1994م وأنجبا ثلاثة أطفال هم: جينفر كاثرين (1996)م، روري جون (1999)م، فيبي أديل (2002)م. وتعيش العائلة في منزل عصري ضخم ومكلف يطل على بحيرة في العاصمة واشنطن. منذ عام 1996م وحتى 2006م حمل بيل غيتس لقب «أغنى رجل في العالم»، فقد قدرت ثروته في عام 1999م بـ100 مليار دولار أمريكي وقد تربع على العرش مرة أخرى عام 2007م.}
        
        ({\bf Translated:} {\it ``Bill Gates married Melinda French in 1994 and they have three children: Jennifer Katherine (1996), Rory John (1999), and Phoebe Adele (2002). The family lives in a huge, expensive modern house overlooking a lake in Washington, DC. From 1996 AD until 2006 AD, Bill Gates held the title of ``the richest man in the world.'' His wealth was estimated in 1999 at 100 billion US dollars, and he ascended to the throne again in 2007 AD.''}\\
        \midrule
        \bf Tabular data & \langar{حصل نتنياهو على درجة البكالوريوس في العلوم.}
        
        {\bf Translated:} {\it ``Netanyahu received a Bachelor of Science degree.''} 
        & Native label: T, Model label: F, Ground truth: T.
        
        {\bf Comment:} Related information is embedded in the table (infobox) which is not processed by Wikipedia's HTML conversion to text, thus not being contained as passages to retrieve.
        
        {\bf Evidence:} Information locates on \langar{المدرسة الأم } (Alma mater) section of the side info box.\\
        \midrule
        \bf Subjective opinion & \langar{براينت كان معروفًا بتفانيه في اللعبة.}
        
        {\bf Translated:} {\it ``Bryant was known for his dedication to the game.''} 
        & Native label: F, Model label: T, Ground truth: T/F.
        
        {\bf Comment:} The statement/fact is subjective, thus debatable. While there is no explicit information that Kobe Bryant was known for his dedication to the game, it could be deductible.\\
        % \midrule
        % \bf Annotation error & \langar{شارك لو تروجليو في العديد من البرامج التلفزيونية.}
        
        % {\bf Translated:}{\it ``Lo Truglio has participated in many TV shows.''} & Native label: F, Model label: T, Ground truth: T.
        
        % {\bf Comment:} He did work on many TV series.
        
        % {\bf Evidence:} There is a list of TV series that Jo Truglio has worked on from his Wikipedia page.\\
        \bottomrule
    \end{tabular}

\caption{Examples from each disagreement category between natives and GPT-4 in Arabic.}
\label{tab:example_disgreement_gpt4_ar}
\end{table*}

\clearpage
\begin{table*}
    \small
    \centering
    \begin{tabular}{|p{0.5\linewidth}|p{0.5\linewidth}|}
        \hline
        \mtc{2}{Original sentence (in English): {\it He then moved to Olympique Alès in 2003 and Stade Brestois in 2004.}} \\
        \hline
        \mtc{2}{- Facts extracted by GPT4 in {\bf English}:}\\
        \mtc{2}{+  He then moved to Olympique Alès.}\\
        \mtc{2}{+  He then moved to Olympique Alès in 2003.}\\
        \mtc{2}{+  He then moved to Stade Brestois.}\\
        \mtc{2}{+  He then moved to Stade Brestois in 2004.}\\
        \mtc{2}{+  Olympique Alès is a team.}\\
        \mtc{2}{+  Stade Brestois is a team.}\\
        \mtc{2}{+  He moved to Olympique Alès and Stade Brestois in consecutive years.}\\
        \hline
        \mtc{2}{- Facts extracted by GPT4 in {\bf Spanish}:}\\
        \mtc{2}{+  Se mudó al Olympique Alès. (En: He then moved to Olympique Alès.)}\\
        \mtc{2}{+  Se mudó al Olympique Alès en 2003. (En: He then moved to Olympique Alès in 2003.)}\\
        \mtc{2}{+  Olympique Alès es un equipo. (En: Olympique Alès is a team.)}\\
        \mtc{2}{+  Luego se mudó al Stade Brestois. (En: He then moved to Stade Brestois.)}\\
        \mtc{2}{+  Se mudó al Stade Brestois en 2004. (En: He moved to Stade Brestois in 2004.)}\\
        \mtc{2}{+  Stade Brestois es un equipo. (En: Stade Brestois is a team.)}\\
        \hline
        \mtc{2}{- Facts extracted by GPT4 in {\bf Arabic}:}\\
        \mtc{2}{+  \langar{ انتقل إلى أولمبيك أليس}. (En: Moved to Olympique Ales)}\\
        \mtc{2}{+   \langar{انتقل إلى أولمبيك أليس في عام 2003.} (En: He moved to Olympique Ales in 2003)}\\
        \mtc{2}{+  \langar{ انتقل بعد ذلك إلى ستاد بريستويس.} (En: He then moved to Stade Brestois.)}\\
        \mtc{2}{+  \langar{ انتقل إلى ستاد بريستويس في عام 2004.} (En: He moved to Stade Brestois in 2004.)}\\
        % \hline
        % \mtc{2}{- Facts extracted by GPT4 in {\bf Bengali}:}\\
        % \mtc{2}{+ অলিম্পিক অ্যালেসে চলে যাওয়ার পর তিনি স্টেড ব্রেস্টয়েসে চলে যান।}\\ 
        % \mtc{2}{\;\;\;\;(En: He moved to Stade Prestois after moving to Olympique Ales) }\\
        % \mtc{2}{+ তিনি অলিম্পিক অ্যালেসে চলে যান। (En: He moved to Olympic Ales.) }\\
        % \mtc{2}{+ তিনি 2003 সালে অলিম্পিক অ্যালেসে চলে যান। (En: He moved to Olympic Ales in 2003.) }\\
        % \mtc{2}{+ তিনি স্টেড ব্রেস্টয়েসে চলে যান। (En: He moved to Stade Brestois.) }\\
        % \mtc{2}{+ তিনি 2004 সালে স্টেড ব্রেস্টয়েসে চলে যান। (En: He moved to Stade Brestois in 2004.) }\\
        \hline
        
    \end{tabular}
\caption{Example of atomic facts extracted by GPT4.}
\label{tab:fact_extraction_gpt4}
\end{table*}
\begin{table*}
    \small
    \centering
    \begin{tabular}{|p{0.5\linewidth}|p{0.5\linewidth}|}
        \hline
        \mtc{2}{Original sentence (in English): {\it He then moved to Olympique Alès in 2003 and Stade Brestois in 2004.}} \\
        \hline
        \mtc{2}{- Facts extracted by GemP in {\bf English}:}\\
        \mtc{2}{+  He moved to Olympique Alès.}\\
        \mtc{2}{+  He moved to Olympique Alès in 2003.}\\
        \mtc{2}{+  He moved to Stade Brestois.}\\
        \mtc{2}{+  He moved to Stade Brestois in 2004.}\\
        \hline
        \mtc{2}{- Facts extracted by GemP in {\bf Spanish}:}\\
        \mtc{2}{+  Se mudó al Olympique Alès. (En: He moved to Olympique Alès.)}\\
        \mtc{2}{+  Se mudó al Olympique Alès en 2003. (En: He moved to Olympique Alès in 2003.)}\\
        \mtc{2}{+  Se mudó al Stade Brestois. (En: He moved to Stade Brestois.)}\\
        \mtc{2}{+  Se mudó al Stade Brestois en 2004. (En: He moved to Stade Brestois in 2004.)}\\
        \hline
        \mtc{2}{- Facts extracted by GemP in {\bf Arabic}:}\\
        \mtc{2}{+ \langar{  انتقل بعد ذلك إلى أولمبيك.} (En: He moved to Olympique)}\\
        \mtc{2}{+  \langar{ انتقل إلى أولمبيك أليس في عام 2003.} (En: He moved to Olympique in 2003)}\\
        \mtc{2}{+  \langar{ انتقل بعد ذلك إلى ستاد بريستويس. }(En: He moved to Stade Prestois.)}\\
        \mtc{2}{+  \langar{ انتقل إلى ستاد بريستويس في عام 2004.} (En: He moved to Stade Prestois in 2004.)}\\
        \hline
        % \mtc{2}{- Facts extracted by GemP in {\bf Bengali}:}\\
        % \mtc{2}{+ তিনি অলিম্পিক অ্যালেসে চলে যান। (En: He moved on to Olympic Ales.) }\\
        % \mtc{2}{+ তিনি 2003 সালে অলিম্পিক অ্যালেসে চলে যান। (En: He moved to Olympic Ales in 2003.) }\\
        % \mtc{2}{+ তিনি স্টেড ব্রেস্টয়েসে চলে যান। (En: He moved to Stade Brestois.) }\\
        % \mtc{2}{+ তিনি 2004 সালে স্টেড ব্রেস্টয়েসে চলে যান। (En: He moved to Stade Brestois in 2004.) }\\
        % \mtc{2}{+ অলিম্পিক অ্যালেসে চলে যাওয়ার পর তিনি স্টেড ব্রেস্টয়েসে চলে যান।}\\
        % \mtc{2}{\;\;\;\;(En: After moving to Olympique Ales, he moved to Stade Brestois.)}\\
        % \mtc{2}{+ 2003 সালে অলিম্পিক অ্যালেসে চলে যাওয়ার পর তিনি 2004 সালে স্টেড ব্রেস্টয়েসে চলে যান।}\\
        % \mtc{2}{\;\;\;\;(En: After moving to Olympique Ales in 2003, he moved to Stade Brestois in 2004.)}\\
        % \hline
        
    \end{tabular}
\caption{Example of atomic facts extracted by GemP.}
\label{tab:fact_extraction_gemini}
\end{table*}
\begin{table*}
    \small
    \centering
    \begin{tabular}{|p{0.5\linewidth}|p{0.5\linewidth}|}
        \hline
        \mtc{2}{Original sentence (in English): {\it He then moved to Olympique Alès in 2003 and Stade Brestois in 2004.}} \\
        \hline
        \mtc{2}{- Facts extracted by GPT3.5 in {\bf English}:}\\
        \mtc{2}{+  He then moved to Olympique Alès.}\\
        \mtc{2}{+  He then moved to Olympique Alès in 2003.}\\
        \mtc{2}{+  He then moved to Stade Brestois.}\\
        \mtc{2}{+  He then moved to Stade Brestois in 2004.}\\
        \hline
        \mtc{2}{- Facts extracted by GPT3.5 in {\bf Spanish}:}\\
        \mtc{2}{+  Se mudó al Olympique Alès. (En: He moved to Olympique Alès.)}\\
        \mtc{2}{+  Se mudó al Olympique Alès en 2003. (En: He moved to Olympique Alès in 2003.)}\\
        \mtc{2}{+  Se mudó al Stade Brestois. (En: He moved to Stade Brestois.)}\\
        \mtc{2}{+  Se mudó al Stade Brestois en 2004. (En: He moved to Stade Brestois in 2004.)}\\
        \hline
        \mtc{2}{- Facts extracted by GPT3.5 in {\bf Arabic}:}\\
        \mtc{2}{+  \langar{انتقل بعد ذلك إلى أولمبيك.} (En: He moved to Olympique Ales)}\\
        \mtc{2}{+  \langar{ انتقل بعد ذلك إلى أولمبيك في عام 2003.} (En: He moved to Olympique Ales in 2003)}\\
        \mtc{2}{+  \langar{ انتقل بعد ذلك إلى ستاد بريستويس.}(En: He moved to Stade Brestois.)}\\
        \mtc{2}{+  \langar{ انتقل إلى ستاد بريستويس في عام 2004.} (En: He moved to Stade Brestois in 2004.)}\\
        \hline
        % \mtc{2}{- Facts extracted by GPT3.5 in {\bf Bengali}:}\\
        % \mtc{2}{+ তিনি অলিম্পিক অ্যালেসে চলে যান। (En: He moved on to Olympic Ales.) }\\
        % \mtc{2}{+ তিনি 2003 সালে অলিম্পিক অ্যালেসে চলে যান। (En: He moved to Olympic Ales in 2003.) }\\
        % \mtc{2}{+ তিনি স্টেড ব্রেস্টয়েসে চলে যান। (En: He moved to Stade Brestois.) }\\
        % \mtc{2}{+ তিনি 2004 সালে স্টেড ব্রেস্টয়েসে চলে যান। (En: He moved to Stade Brestois in 2004.) }\\
        % \hline
        
    \end{tabular}
\caption{Example of atomic facts extracted by GPT3.5.}
\label{tab:fact_extraction_gpt3.5}
\end{table*}

\begin{table*}
    \small
    \centering
    \begin{tabular}{|p{0.5\linewidth}|p{0.5\linewidth}|}
        \hline
        \mtc{2}{Original sentence (in English): {\it He then moved to Olympique Alès in 2003 and Stade Brestois in 2004.}} \\
        \hline
        \mtc{2}{- Facts extracted by Mistral Instruct in {\bf English}:}\\
        \mtc{2}{+  He then moved to Olympique Alès.}\\
        \mtc{2}{+  He then moved to Olympique Alès in 2003.}\\
        \mtc{2}{+  He then moved to Stade Brestois.}\\
        \mtc{2}{+  He then moved to Stade Brestois in 2004.}\\
        \hline
        \mtc{2}{- Facts extracted by Mistral in {\bf Spanish}:}\\
        \mtc{2}{+  Se mudó al Olympique Alès. (En: He moved to Olympique Alès.)}\\
        \mtc{2}{+  Se mudó al Olympique Alès en 2003. (En: He moved to Olympique Alès in 2003.)}\\
        \mtc{2}{+  Se mudó al Stade Brestois. (En: He moved to Stade Brestois.)}\\
        \mtc{2}{+  Se mudó al Stade Brestois en 2004. (En: He moved to Stade Brestois in 2004.)}\\
        \hline
        \mtc{2}{- Facts extracted by Mistral Instruct in {\bf Arabic}:}\\
        \mtc{2}{+  \langar{ انتقل بعد ذلك.} (En: \red{He moved: INCOMPLETE})}\\
        \mtc{2}{+  \langar{ انتقل بعد ذلك إلى أولمبيك.} (En: He moved to Olympique)}\\
        \mtc{2}{+  \langar{ انتقل بعد ذلك إلى أولمبيك في عام 2003.} (En: He moved to Olympique in 2003)}\\
        \mtc{2}{+  \langar{ ستاد بريستويس.} (En: \red{Stade Prestois - INCOMPLETE})}\\
        \mtc{2}{+  \langar{ ستاد بريستويس في عام 2004.} (En: \red{Stade Prestois in 2004 - INCOMPLETE})}\\
        \mtc{2}{+   \langar{انتقل بعد ذلك إلى أولمبيك في عام 2003 وستاد بريستويس في عام 2004. }}\\
        \mtc{2}{\;\;\;\;(En: He then moved to Olympique in 2003 and Stade Prestois in 2004.)}\\
        \hline
        % \mtc{2}{- Facts extracted by Mistral Instruct in {\bf Bengali}:}\\
        % \mtc{2}{+ তিনি 2003 সালে অলিম্পিক অ্যালেসে চলে যান। (En: He moved to Olympic Ales in 2003.) }\\
        % \mtc{2}{+ তিনি 2004 সালে স্টেড ব্রেস্টয়েসে চলে যান। (En: He moved to Stade Brestois in 2004.) }\\
        % \mtc{2}{+ অলিম্পিক অ্যালেসে এবং স্টেড ব্রেস্টয়েসে দুটো বিদেশী স্পোর্টস স্টেডিও। }\\
        % \mtc{2}{\;\;\;\;(\red{En: Two foreign sports stadiums in Olympique Ales and Stade Brestois. - NOT GROUNDED}) }\\
        % \mtc{2}{+ তিনি দুটো বিদেশী স্পোর্টস স্টেডিও চলে যান। }\\
        % \mtc{2}{(\red{En: He also visited two foreign sports stadiums. - NOT GROUNDED})}\\
        % \mtc{2}{+ তিনি 2003 সালে অলিম্পিক অ্যালেসে এবং 2004 সালে স্টেড ব্রেস্টয়েসে চলে যান। }\\
        % \mtc{2}{\;\;\;\;(\red{En: He moved to Olympique Ales in 2003 and Stade Brestois in 2004. - NOT ATOMIC}) }\\
        % \hline
        
    \end{tabular}
\caption{Example of atomic facts extracted by Mistral-Instruct. }
\label{tab:fact_extraction_mistral}
\end{table*}

        % \hline
        % \mtc{2}{- Facts extracted by Mistral in {\bf Arabic}:}\\
        % \mtc{2}{+   انتقل بعد ذلك. (En: \red{He moved: INCOMPLETE})}\\
        % \mtc{2}{+   انتقل بعد ذلك إلى أولمبيك. (En: He moved to Olympique)}\\
        % \mtc{2}{+   انتقل بعد ذلك إلى أولمبيك في عام 2003. (En: He moved to Olympique in 2003)}\\
        % \mtc{2}{+   ستاد بريستويس. (En: \red{Stade Prestois - INCOMPLETE})}\\
        % \mtc{2}{+   ستاد بريستويس في عام 2004. (En: \red{Stade Prestois in 2004 - INCOMPLETE})}\\
        % \mtc{2}{+   انتقل بعد ذلك إلى أولمبيك في عام 2003 وستاد بريستويس في عام 2004. }\\
        % \mtc{2}{\;\;\;\;(En: \red{He then moved to Olympique in 2003 and Stade Prestois in 2004. - NOT ATOMIC})}\\
        % \hline
        % \mtc{2}{- Facts extracted by Mistral in {\bf Bengali}:}\\
        % \mtc{2}{+ তিনি 2003 সালে অলিম্পিক অ্যালেসে চলে যান। (En: He moved to Olympic Ales in 2003.) }\\
        % \mtc{2}{+ তিনি 2004 সালে স্টেড ব্রেস্টয়েসে চলে যান। (En: He moved to Stade Brestois in 2004.) }\\
        % \mtc{2}{+ অলিম্পিক অ্যালেসে এবং স্টেড ব্রেস্টয়েসে দুটো বিদেশী স্পোর্টস স্টেডিও। }\\
        % \mtc{2}{\;\;\;\;(\red{En: Two foreign sports stadiums in Olympique Ales and Stade Brestois. - NOT GROUNDED}) }\\
        % \mtc{2}{+ তিনি দুটো বিদেশী স্পোর্টস স্টেডিও চলে যান। (\red{En: He also visited two foreign sports stadiums. - NOT GROUNDED}) }\\
        % \mtc{2}{+ তিনি 2003 সালে অলিম্পিক অ্যালেসে এবং 2004 সালে স্টেড ব্রেস্টয়েসে চলে যান। }\\
        % \mtc{2}{\;\;\;\;(\red{En: He moved to Olympique Ales in 2003 and Stade Brestois in 2004. - NOT ATOMIC}) }\\
        % \hline

\begin{table*}
    \small
    \centering
    \begin{tabular}{|p{0.5\linewidth}|p{0.5\linewidth}|}
        \hline
        \mtc{2}{Original sentence (in English): {\it He then moved to Olympique Alès in 2003 and Stade Brestois in 2004.}} \\
        \hline
        \mtc{2}{- Facts extracted by Llama-2 Chat in {\bf English}:}\\
        \mtc{2}{+  He moved to Olympique Alès.}\\
        \mtc{2}{+  He moved to Olympique Alès in 2003.}\\
        \mtc{2}{+  He moved to Stade Brestois.}\\
        \mtc{2}{+  He moved to Stade Brestois in 2004.}\\
        \hline
        \mtc{2}{- Facts extracted by Llama-2 Chat in {\bf Spanish}:}\\
        \mtc{2}{+  Luego se mudó. (En: Then he moved.)}\\
        \mtc{2}{+  Se mudó al Olympique Alès. (En: He moved to Olympique Alès.)}\\
        \mtc{2}{+  Se mudó al Stade Brestois. (En: He moved to Stade Brestois.)}\\
        \mtc{2}{\;\;\;\;\red{Insufficient covering information from the original sentence, years of movement}}\\
        \hline
        \mtc{2}{- Facts extracted by Llama-2 Chat in {\bf Arabic}:}\\
        \mtc{2}{+\langar{ ثم انتقل.} (En: \red{Then move on: INCOMPLETE})}\\
        \mtc{2}{+\langar{ انتقل بعد ذلك.} (En: \red{Go then: INCOMPLETE})}\\
        \mtc{2}{+\langar{ إلى أولمبيك أليس.} (En: \red{To Olympique Alice: INCOMPLETE})}\\
        \mtc{2}{+\langar{ إلى أولمبيك أليس في عام 2003.} (En: \red{To Olympique Alice in 2003: INCOMPLETE})}\\
        \mtc{2}{+\langar{ ستاد بريستويس.} (En: \red{Prestois Stadium: INCOMPLETE})}\\
        \mtc{2}{+\langar{ ستاد بريستويس في عام 2004.}  (En: \red{Prestois Stadium in 2004: INCOMPLETE})}\\
        \hline
        \mtc{2}{- Facts extracted by Llama-2 Chat in {\bf Bengali}:}\\
        \mtc{2}{\;\;\;\;\red{Does not have sufficient tokens in Bengali (text with full of UNK tokens)}}\\
        \hline
        
    \end{tabular}
\caption{Example of atomic facts extracted by Llama-2 Chat. }
\label{tab:fact_extraction_llama2}
\end{table*}

\begin{table*}
    \centering
    \begin{tabular}{|p{0.5\linewidth}|p{0.5\linewidth}|}
        \hline
        \mtc{2}{Original sentence (in English): {\it He then moved to Olympique Alès in 2003 and Stade Brestois in 2004.}} \\
        \hline
        \mtc{2}{- Facts extracted by Gemma Instruct in {\bf English}:}\\
        \mtc{2}{+  He moved to Olympique Alès.}\\
        \mtc{2}{+  He moved to Olympique Alès in 2003.}\\
        \mtc{2}{+  He moved to Stade Brestois.}\\
        \mtc{2}{+  He moved to Stade Brestois in 2004.}\\
        \hline
        \mtc{2}{- Facts extracted by Gemma Instruct in {\bf Spanish}:}\\
        \mtc{2}{+  Se mudó al Olympique Alès. (En: He moved to Olympique Alès.)}\\
        \mtc{2}{+  Se mudó al Olympique Alès en 2003. (En: He moved to Olympique Alès in 2003.)}\\
        \mtc{2}{+  Se mudó al Stade Brestois. (En: He moved to Stade Brestois.)}\\
        \mtc{2}{+  Se mudó al Stade Brestois en 2004. (En: He moved to Stade Brestois in 2004.)}\\
        \hline
        \mtc{2}{- Facts extracted by Gemma Instruct in {\bf Arabic}:}\\
        \mtc{2}{+\langar{ انتقل إلى أولمبيك.} (En: \red{Moved to Olympic: DUPLICATED})}\\
        \mtc{2}{+\langar{ انتقل إلى أولمبيك.} (En: \red{Moved to Olympic: DUPLICATED})}\\
        \mtc{2}{+\langar{ انتقل إلى أولمبيك.} (En: \red{Moved to Olympic: DUPLICATED})}\\
        \mtc{2}{+\langar{ انتقل إلى أولمبيك.} (En: \red{Moved to Olympic: DUPLICATED})}\\
        \hline
        % \mtc{2}{- Facts extracted by Gemma Instruct in {\bf Bengali}:}\\
        % \mtc{2}{+ তিনি স্টেড চলে যান। (En: He went to Stade.)}\\
        % \mtc{2}{+ তিনি স্টেড চলে যান। (En: \red{He went to Stade: DUPLICATED})}\\
        % \mtc{2}{+ তিনি স্টেড চলে যান। (En: \red{He went to Stade: DUPLICATED})}\\
        % \mtc{2}{+ তিনি স্টেড চলে যান। (En: \red{He went to Stade: DUPLICATED})}\\
        % \hline
        
    \end{tabular}
\caption{Example of atomic facts extracted by Gemma-7B-Instruct. }
\label{tab:fact_extraction_gemma_instr}
\end{table*}
\begin{table*}
    \small
    \centering
    \begin{tabular}{|p{0.5\linewidth}|p{0.5\linewidth}|}
        \hline
        \mtc{2}{Original sentence (in English): {\it He then moved to Olympique Alès in 2003 and Stade Brestois in 2004.}} \\
        \hline
        \mtc{2}{- Facts extracted by Finetuned Gemma in {\bf English}:}\\
        \mtc{2}{+  He moved to Olympique Alès.}\\
        \mtc{2}{+  He moved to Olympique Alès in 2003.}\\
        \mtc{2}{+  He moved to Stade Brestois.}\\
        \mtc{2}{+  He moved to Stade Brestois in 2004.}\\
        \hline
        \mtc{2}{- Facts extracted by Finetuned Gemma in {\bf Spanish}:}\\
        \mtc{2}{+  Se mudó al Olympique Alès. (En: He moved to Olympique Alès.)}\\
        \mtc{2}{+  Se mudó al Olympique Alès en 2003. (En: He moved to Olympique Alès in 2003.)}\\
        \mtc{2}{+  Se mudó al Stade Brestois. (En: He moved to Stade Brestois.)}\\
        \mtc{2}{+  Se mudó al Stade Brestois en 2004. (En: He moved to Stade Brestois in 2004.)}\\
        \hline
        \mtc{2}{- Facts extracted by Finetuned Gemma in {\bf Arabic}:}\\
        \mtc{2}{+ \langar{  انتقل بعد ذلك إلى أولمبيك.} (En: He moved to Olympique)}\\
        \mtc{2}{+  \langar{ انتقل إلى أولمبيك أليس في عام 2003.} (En: He moved to Olympique in 2003)}\\
        \mtc{2}{+  \langar{ انتقل بعد ذلك إلى ستاد بريستويس.} (En: He moved to Stade Prestois.)}\\
        \mtc{2}{+  \langar{ انتقل إلى ستاد بريستويس في عام 2004.} (En: He moved to Stade Prestois in 2004.)}\\
        \hline
        % \mtc{2}{- Facts extracted by Finetuned Gemma in {\bf Bengali}:}\\
        % \mtc{2}{+ তিনি অলিম্পিক অ্যালেসে চলে যান। (En: He moved to Olympic Ales.) }\\
        % \mtc{2}{+ তিনি 2003 সালে অলিম্পিক অ্যালেসে চলে যান। (En: He moved to Olympic Ales in 2003.) }\\
        % \mtc{2}{+ তিনি স্টেড ব্রেস্টয়েসে চলে যান। (En: He moved to Stade Prestois.) }\\
        % \mtc{2}{+ তিনি 2004 সালে স্টেড ব্রেস্টয়েসে চলে যান। (En: He moved to Stade Prestois in 2004.) }\\
        % \hline
        
    \end{tabular}
\caption{Example of atomic facts extracted by Finetuned Gemma. }
\label{tab:fact_extraction_ft_gemma}
\end{table*}

\begin{table*}[!h]
    \small
    \centering
    \begin{tabular}{cp{0.2\linewidth}p{0.7\linewidth}}
    \toprule
        
        & \bf Fact & \bf Google API Texts\\
        \midrule

        \mtr{6}{\bf es}
        & {\it ``Spivak ha escrito sobre la traducción''} 
        
        $\rightarrow$ {\it ``Spivak has written about translation''} 
        % & Golden label: 
        
        % {\bf Supported }
        
        % w/o Google API: 
        
        % {\bf Not Supported}
        
        % w/ Google API:

        % {\bf Supported}
        &
        
        {\it ``Gayatri Chakravorty Spivak is an Indian scholar, literary theorist, and feminist critic. She has made a significant statement on “the politics of translation” ... Living Translation performs the invaluable service of gathering for the first time Gayatri Chakravorty Spivak's wide-ranging writings on translation.''}
        
        {\bf Comment:} Gayatri Chakravorty Spivak's Wikipedia page has no information about her research on the impact of translation. Yet, using Google Query API returns information about her two books on the topic ("The Politics of Transaltion" and "Living Translation").\\
        \midrule
        \mtr{6}{\bf ar} & \langar{خصخص العديد من الشركات الحكومية.}
        
        $\rightarrow$ {\it ``He privatized many state companies.''}
        % & Golden label: 
        
        % {\bf Supported }
        
        % w/o Google API: 
        
        % {\bf Not Supported}
        
        % w/ Google API:

        % {\bf Supported}
        & 
        
        \langar{ويستفاد من معطيات التقرير أن أكبر عمليات الخوصصة في تاريخ المغرب قد تمت خلال حكومتي عبد الرحمن اليوسفي وادريس جطو، وقد أدرت أموالاً كبيرة}
        
        $\rightarrow$ {\it ``It is clear from the report’s data that the largest privatization operations in the history of Morocco took place during the governments of Abderrahmane Youssoufi and Idriss Jettou, and they generated large sums of money.''}
        
        {\bf Comment:} The additional information supports the fact about privatization initiatives under Idriss Jettou's term, whereas the Wikipedia page has no related information about it.\\

        \midrule
        % \mtr{6}{\bf bn} & "তিনি ইরানি পরমাণু কর্মসূচির বিরোধীতা করেছেন।'" 
        
        % $\rightarrow$ {\it ``He opposes the Iranian nuclear program.''}
        
        % 'sent': 'He opposes the Iranian nuclear program and calls for military action against Iran.”) 
        % & Golden label: 
        
        % {\bf Supported }
        
        % w/o Google API: 
        
        % {\bf Not Supported}
        
        % w/ Google API:

        % {\bf Supported}
        % & 
        
        % {\it ``Israeli Prime Minister Benjamin Netanyahu has spent the past three decades sounding the alarm about Iran's nuclear program and threatening...Benjamin Netanyahu said Tuesday that he would change Israel's strategy toward countering Iranian nuclear ambitions.''}
        
        % {\bf Comment:} The additional information supports the fact about Israeli Prime Minister Benjamin Netanyahu's opposition to the Iranian nuclear program, whereas the Wikipedia page has no related information.\\
        
        % \bottomrule
    \end{tabular}
\caption{Examples demonstrate that using Google Query API provides additional information to the scorer, GemP, leading to accurate fact labeling. The additional information from the examples is not present on Wikipedia pages and has been manually validated as correct.}
\label{tab:googleapi_example}
\end{table*}

\begin{table*}[!h]
    \small
    \centering
    \begin{tabular}{cp{0.2\linewidth}p{0.7\linewidth}}
    \toprule
         & \bf Fact \& Labels & \bf GPT4 Additional Texts\\
        \midrule
        \bf es & {\it Implementó políticas radicals.}
        
        $\rightarrow$ {\it ``He implemented radical policies.''}
        % & Golden:

        % {\bf Supported}
        
        % w/o GPT4's IK:

        % {\bf Not Supported}
        
        % w/ GPT4's IK:

        % {\bf Supported}

        & 

        {\bf Query:} {\it ``¿Qué tipo de políticas implementó Hafizullah Amín?''}
        
        $\rightarrow$ {\it ``What kind of policies did Hafizullah Amin implement?''}
        
        {\bf GPT-4 Texts:} {\it ``Durante su tiempo en el poder, Amín se desempeñó como primer ministro y luego, brevemente, como presidente de Afganistán desde septiembre hasta diciembre de 1979. Sus políticas se centraron en la implementación de una agenda socialista radical, que incluía reformas agrarias destinadas a abolir las tradicionales tenencias feudales de tierra y redistribuir las tierras a los campesinos sin tierra.''}
        
        $\rightarrow$ {\it ``During his time in power, Amin served as prime minister and then briefly as president of Afghanistan from September to December 1979. His policies focused on the implementation of a radical socialist agenda, which included land reforms aimed at abolishing the traditional feudal land tenures and redistributing land to landless peasants.''}
        
        {\bf Comment:} The additional information generated by GPT4 supports the fact about radical reforms by Hafizullah Amin, whereas the Wikipedia page has no related information about it. The generated information is manually confirmed to be correct\\
        
        \midrule
        \bf ar & \langar{بدأ عرض "The Apprentice" في عام 2004.}
        
        $\rightarrow$ {\it ``The Apprentice" began airing in 2004.''}
        
        % & Golden:

        % {\bf Supported}
        
        % w/o GPT4's IK:

        % {\bf Not Supported}
        
        % w/ GPT4's IK:

        % {\bf Supported}

        & 
        {\bf Query:} \langar{متى بدأ عرض برنامج "The Apprentice" الذي كان دونالد ترامب يقدمه؟}
        
        $\rightarrow$ {\it ``When did Donald Trump's show "The Apprentice" start airing?''}
        
        {\bf GPT-4 Texts:} \langar{برنامج "The Apprentice" هو برنامج تلفزيوني أمريكي من نوع تلفزيون الواقع، بدأ عرضه في الولايات المتحدة. ظهر البرنامج لأول مرة في 8 يناير 2004 على شبكة NBC.}
        
        $\rightarrow$ {\it ``The Apprentice" is an American reality television show that began airing in the United States. The program debuted on January 8, 2004 on NBC.''}
        
        {\bf Comment:} The additional information generated by GPT4 supports the fact about the airing time of "The Apprentice", whereas the Wikipedia page has no related information about it. The generated information is manually confirmed to be correct.\\

        \midrule

        \bottomrule
    \end{tabular}
\caption{Examples demonstrate that using GPT-4 as a knowledge generator provides additional information to the scorer, GemP, leading to accurate fact labeling. The additional information from the examples is not present on Wikipedia pages and has been manually validated as correct.}
\label{tab:gpt4ks_example}
\end{table*}

\end{document}